\documentclass[15,a4paper]{article}
\usepackage[margin=25mm]{geometry}
\usepackage{amsmath}
\usepackage{amsfonts}
\usepackage{amssymb}
\usepackage{graphicx}
\usepackage{verbatim}
\usepackage{graphicx}

\usepackage{tabularx}
\usepackage{algorithm}
\usepackage{watermark}
\usepackage{hyperref}

\usepackage{verbatim}
\usepackage{array}
\usepackage{multirow}
\usepackage{pifont}
\usepackage{amssymb}
\usepackage{mathtext}
\usepackage{booktabs,siunitx} % for professional tables
\usepackage{mathtools}
\usepackage{fontenc}
\usepackage{microtype}
\usepackage{xcolor}
\usepackage{colortbl}
\usepackage{textcomp}
\usepackage{url}
\usepackage{subfigure}
\usepackage{appendix}

\usepackage{enumerate}

\usepackage{algorithm}
\usepackage{array}
\usepackage{eqparbox}
\usepackage{array}
\usepackage{eqparbox}
\usepackage{flushend}

\usepackage[normalem]{ulem}
\usepackage{algpseudocode}
\usepackage{float}
\floatstyle{plaintop}
\restylefloat{table}
\immediate\write18{texcount -tex -sum  \jobname.tex > \jobname.wordcount.tex}

\providecommand{\keywords}[1]
{
  \small	
  \textbf{\textit{Keywords---}} #1
}

\title{Light-weight Document Image Cleanup using Perceptual Loss}
\author{  Soumyadeep Dey, Pratik Jawanpuria\\
       Microsoft India \\
       \{soumyadeep.dey,pratik.jawanpuria\}@microsoft.com \\
}
\date{} % Comment this line to show today's date

\begin{document}
\maketitle

\begin{abstract}
%The abstract should briefly summarize the contents of the paper in 150--250 words.
Smartphones have enabled effortless capturing and sharing of documents in digital form. The documents, however, often undergo various types of degradation due to aging, stains, or shortcoming of capturing environment such as shadow, non-uniform lighting, etc., which reduces the comprehensibility of the document images. 
In this work, we consider the problem of document image cleanup on embedded applications such as smartphone apps, which usually have memory, energy, and latency limitations due to the device and/or for best human user experience. 
We propose a light-weight encoder decoder based convolutional neural network architecture for removing the noisy elements from document images. 
%To restrict the network size, we employ only a few residual blocks and skip connections. 
%In addition, our loss function incorporates the perceptual loss for knowledge transfer from pre-trained deep CNN network. 
To compensate for generalization performance with a low network capacity, we incorporate the perceptual loss for knowledge transfer from pre-trained deep CNN network in our loss function. 
% To restrict the network size, we incorporates the perceptual loss for knowledge transfer from pre-trained deep CNN network in our loss function. 
% With the widespread use of memory and energy constrained devices like smartphones, it has become essential to have an efficient light-weight document cleanup technique for better readability and user experience. 
% In this work, we have proposed a light-weight encoder-decoder based convolutional neural network (CNN) architecture for the document image cleanup. To learn the complex noise function in a light-weight deep network, we utilize the concept of perceptual loss-based transfer learning technique. 
%In terms of model size on device, our models are 65-1090 times smaller than existing state-of-the-art document enhancement models, while being 1.5-18 times faster in inference stage.
In terms of the number of parameters and product-sum operations, our models are 65-1030 and 3-27 times, respectively, smaller than existing state-of-the-art document enhancement models. 
Overall, the proposed models offer a favorable resource versus accuracy trade-off and we empirically illustrate the efficacy of our approach on several real-world benchmark datasets. 
%Empirical results on several real-world benchmark datasets illustrate the efficacy of our approach. 
%We demonstrate the effectiveness of the proposed method by showcasing its performance on various real-world benchmark datasets. 

%\alertbyPJ{The title should emphasize on mobile/embedded-system friendly scenario.} 
\end{abstract} \hspace{10pt}

%TC:ignore

\section{Introduction}
\label{sec:intro}
\begin{figure}[t]\centering
\begin{tabular}{@{}c@{\ }c@{\ }c@{\ }c@{}}
\fbox{\includegraphics[width=.22\textwidth]{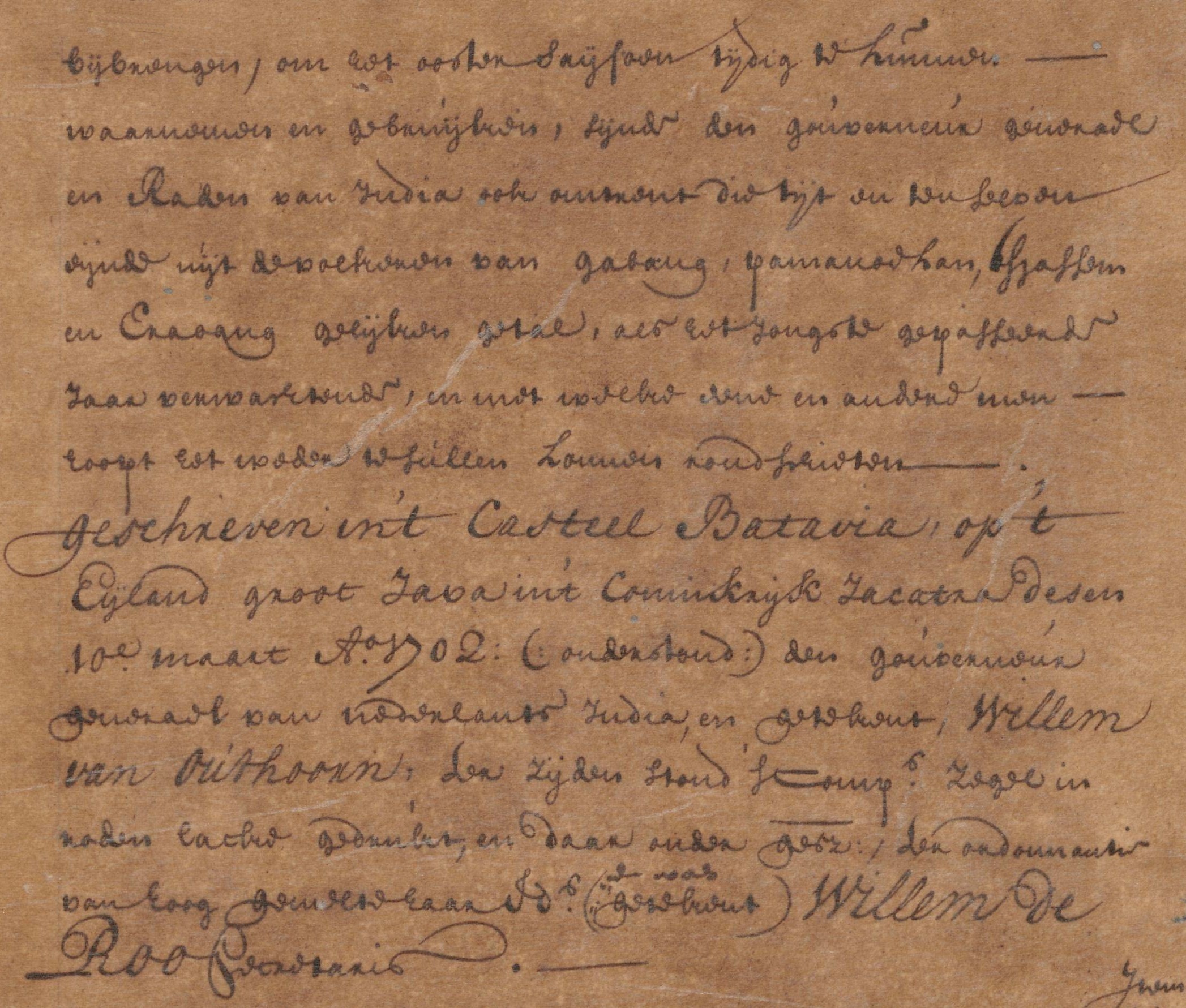}}&
\fbox{\includegraphics[width=.22\textwidth]{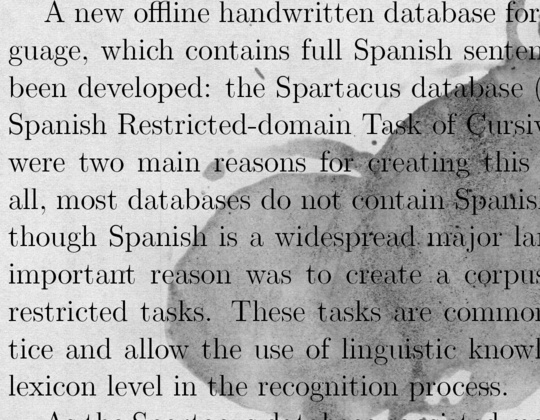}}&
\fbox{\includegraphics[width=.22\textwidth]{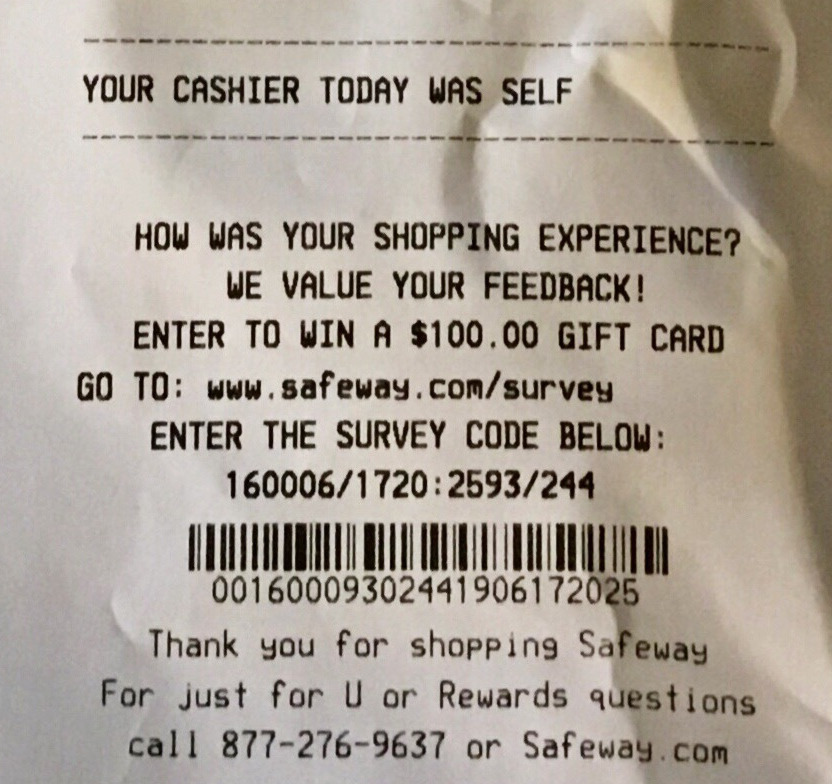}}&
\fbox{\includegraphics[width=.22\textwidth]{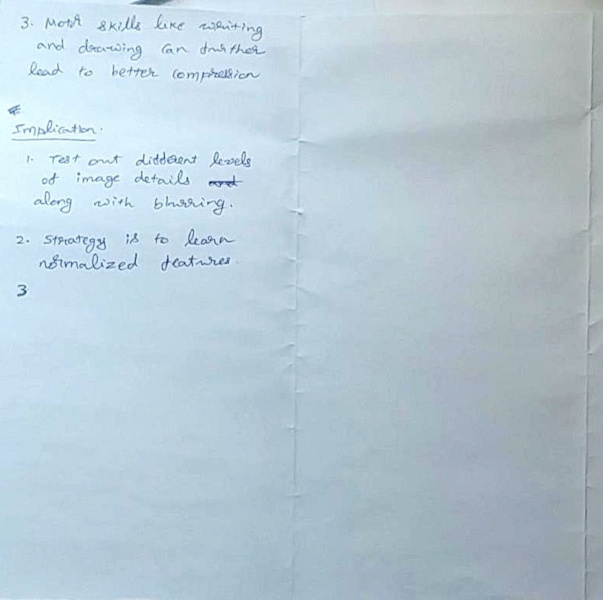}}\\
\multicolumn{2}{c}{(a)} & \multicolumn{2}{c}{(b)} \\
\end{tabular}
\caption{Typical examples of noisy document images; (a) real world degraded images from various datasets \cite{Dua:2019,DIBCO17,NoisyOffice}, (b) noisy real world images captured via mobile devices. }
\label{fig:noisy_example}
\end{figure}

The smartphone camera have simplified the capture of various physical  documents in digital form. The ease of share of digital documents (e.g., via messaging/networking apps) have made them a popular source of information dissemination. However, readability of such digitized documents is hampered when the (original) physical document is degraded. For instance, the physical document may contain extraneous elements like stains, wrinkles, ink spills, or can undergo degradation over time. As a result, while scanning such documents (e.g., via a flat-bed scanner), these elements also get incorporated into the document image. In case of capturing document images via mobile cameras, the images are prone to being impacted by shadow, non-uniform lighting, light from multiple sources, light source occlusion, etc. Such \textit{noisy} elements not only effects the comprehensibility of the corresponding digitized document to the human readers, it may also break down the automatic (document-image) processing/understanding pipeline in various applications (e.g., OCR, bar code reading, form detection, table detection, etc). Few instances of \textit{noisy} document images are shown in Fig.~\ref{fig:noisy_example}. 

Given an input noisy document image, the aim of document image cleanup is to improve its readability and visibility by removing the noisy elements. %, where foreground pixels are preserved and enhanced, and background is made uniform. 
While general (natural scene) image restoration has been traditionally explored by the computer vision community, recent works have also focused on developing cleanup techniques for document images depending on the type of noise and document-class. 
These include foreground background separation \cite{adotsu_pr12,binary_fcmean_icdar19,Sauvola00adaptivedocument}, differential fading problem \cite{Liu_Binary_ICDAR17,sayed10}, removal of shadow/smear/strain \cite{BEDSR-Net_CVPR20,Liu_Binary_ICDAR17,Water-Filling_ACCV18,Tensmeyer_Binary_ICDAR17,valizadeh_binarization,shadow_icip2019,shadow_icassp20}, and handling ink bleed \cite{silva08,Tensmeyer_Binary_ICDAR17}, etc. 

%\begin{figure}[h!]\centering
%\begin{tabular}{@{}c@{\ }c@{\ }c@{}}
%\fbox{\includegraphics[width=.31\textwidth]{./images/example_show/2_in_2.jpeg}}&
%\fbox{\includegraphics[width=.31\textwidth]{./images/example_show/2_in.jpeg}}&
%\fbox{\includegraphics[width=.31\textwidth]{./images/example_show/image_180_in.jpeg}}\\
%\multicolumn{3}{c}{(a)} \\
%(a)&(b)&(c)
%\fbox{\includegraphics[width=.31\textwidth]{./images/example_show/image_2.jpeg}}&
%\fbox{\includegraphics[width=.31\textwidth]{./images/example_show/image-419.jpeg}}&
%\fbox{\includegraphics[width=.31\textwidth]{./images/example_show/image_1.jpeg}}\\
%\multicolumn{3}{c}{(b)} \\
%\end{tabular}
%\caption{Typical examples of noisy document images; (a) real world degraded images from various datasets \cite{DIBCO18,DIBCO17,NoisyOffice,Dua:2019}, (b) noisy real world images captured via mobile devices. }
%\label{fig:noisy_example}
%\end{figure}

%\alertbyPJ{A popular framework for document image clean-up is background foreground separation~\cite{pg_seg_book} (we should give appropriate citations here - preferably of some survey/literature review). Binarization is a technique to segment foreground from the background pixels. }

Recent works~\cite{gangeh2019document,illu_cor2019,Skip-Connected_ICPR18} view document cleanup as an image to image translation problem, modeled using deep networks. A general direction of research has been to explore deeper and more complicated networks in order to achieve better accuracy~\cite{DEGAN_TPAMI20,Skip-Connected_ICPR18}. However, such deep networks often require high computational resources, which is beyond many mobile and embedded applications on a computationally limited platform. Deeper networks also usually entail a higher inference time (latency), which document image processing mobile apps such as Adobe Lens, CamScanner, Microsoft Office Lens, etc., aim to minimize for best human user experience. 
%networks  more number of parameters results in increase in size and there after it become more difficult to run inference on memory constrained devices like mobiles. 
%In  many  real  world  applications  such  as  robotics,self-driving car and augmented reality, the recognition tasks need to be carried out in a timely fashion on a computation-ally limited platform. 

% An image transformation network is a convolutional neural network (CNN) parametrized  by weights $W$, which transforms an input image $y$ into a output image $\hat{y}$ using a mapping function $\hat{y}=\mathit{F}_{W}(y)$. Image transformation networks are trained to minimize a loss function. 
% The deep neural network architectures discussed above are deeper in nature.  
% Deep networks have more number of parameters which helps it to learn complex functions~\cite{DEGAN_TPAMI20,Skip-Connected_ICPR18}. \alertbyPJ{Can we improve upon the contrast between `classical/traditional' methods and the newer deep learning based methods?} 

In this work, we propose a light-weight encoder-decoder based convolutional neural network (CNN) with skip-connections for cleaning up document images. 
Focusing on memory constrained mobile and embedded devices, we design a light-weight deep network architecture. 
It should be noted that light-weight deep network architecture usually costs generalization performance when compared with deeper networks. 
Hence, in order to obtain a healthy interplay between resource/latency and accuracy, we propose to employ perceptual loss function (instead of the more popular per-pixel loss function) for document image cleanup. 
The perceptual loss function~\cite{Johnson2016Perceptual} enables transfer learning by comparing high-level representation of images, obtained from a pre-trained CNNs (e.g., trained on image classification tasks). 
%In order to learn such a complex noise function in a shallow network we proposes to use of perceptual loss~\cite{Johnson2016Perceptual} based transfer learning. 
% Earlier perceptual loss has been used for illumination correction of document images in~\cite{illu_cor2019}. 
% Here we have shown the usefulness of perceptual loss for learning complex noise function in a shallow network. 
% We show the effectiveness of the proposed network on various document cleanup tasks like binarization, and overall document enhancement for both color and gray scale images. 
We empirically show the effectiveness of the proposed network on several real-world benchmark datasets. 
%\alertbyPJ{We should emphasis more on the importance of mobile friendly small models.} 

The outline of the paper is as follows. 
We discuss the existing literature in Section~\ref{sec:RW}. In Section~\ref{sec:method}, we detail our methodology. 
The empirical results are presented in Section~\ref{sec:result}, while Section~\ref{sec:conclusion} concludes the paper.

\section{Related work}
\label{sec:RW}
In this section, we briefly discuss existing approaches that aim to recover/enhance images of degraded documents via techniques involving binarization, and illumination/shadow correction, and deblurring, among others.

%\begin{figure}[h!]
 %\centering
 %\includegraphics[width=.75\textwidth]{./images/Doc_CleanUp.png}
 % LayoutTypes.pdf: 595x842 pixel, 72dpi, 20.99x29.70 cm, bb=0 0 595 842
% \caption{Document clean-up}
 %\label{fig:cleanup}
%\end{figure}

%This work targeted to clean a document image by addressing various type of document image degradation such as, shadow, bleed-through, ageing, illumination correction, folds, crumples, etc. 

% \subsection{Document image binarization}
\noindent\textbf{Document image binarization}: 
A popular framework for document image cleanup is background foreground separation~\cite{pg_seg_book}, where the foreground pixels are preserved and enhanced and the background is made uniform. Binarization is a technique to segment foreground from the background pixels. 
Analytical techniques for document image binarization involve segmenting the foreground pixels and background pixels based on some thresholding. Traditional image binarization technique such as~\cite{otsu} compute a global threshold assuming that the pixel intensity distribution follows a bi-modal histogram. As estimating such thresholds may difficult for degraded document images, Moghaddam and Cheriet, in~\cite{adotsu_pr12}, proposed an adaptive generalization of the Otsu's method~\cite{otsu} for document image binarization. 
In~\cite{Sauvola00adaptivedocument}, Sauvola and Pietikäinen proposed a local adaptive thresholding method for the image binarization task. 
%Sauvola and Pietikäinen proposed an adaptive thresholding method for document image binarization in~\cite{Sauvola00adaptivedocument}. 
To improve Sauvola's algorithm's performance in low contrast setting, Lazzara and Geraud developed its multi-scale generalization in~\cite{lazzara_IJDAR14}. 
% The Sauvola's algorithm computes a local threshold for each pixel, allowing it to enhance images having certain types of degradations but unable to perform well in low contrast setting. 
% To overcome the said limitation, Lazzara and Geraud proposed a multi-scale version  of Sauvola's algorithm in~\cite{lazzara_IJDAR14}. 
% The above techniques use pixel intensity based information to perform local/global threshold to obtain binary image. 
Recent works on document image binarization have also explored techniques based on conditional random fields~\cite{peng_binary_icdar13}, fuzzy C-means clustering~\cite{binary_fcmean_icdar19}, robust regression~\cite{Vo_bin_PR18}, and maximum entropy classification~\cite{Liu_Binary_ICDAR17}. 
Deep convolutional neural networks (CNNs) have become all-pervasive in computer vision ever since AlexNet~\cite{krizhevsky12a} won the ILSVRC~2012 ImageNet Challenge~\cite{russakovsky15a}.  
%With the increase of deep learning based technique for various computer vision problems, researchers came up with various deep learning based approaches for the document image binarization. 
Tensmeyer and Martinez~\cite{Tensmeyer_Binary_ICDAR17} posed document image binarization as a pixel classification problem and developed a fully connected convolution network for it. 
An encoder-decoder network was proposed in~\cite{deep_otsu_pr19} to estimate the background of a document image. Then, Otsu's global thresholding technique~\cite{otsu} is used to obtain a binarized image with uniform background. 
%An encoder-decoder based network was proposed by He and Schomaker for document enhancement and binarization~\cite{deep_otsu_pr19}. In this method, the network is first used to estimate the background and generate uniform background image. Then Otsu's global thresholding technique~\cite{otsu} is used to produce the binary image from the uniform background image. 
Afzal~\textit{et~al.}~\cite{afzal_LSTM_HiP15} employed a long short-term memory (LSTM) network to classify each pixel as background and foreground by considering images to be a two-dimensional sequenec of pixels. 
%Afzal~\textit{et~al.}~\cite{afzal_LSTM_HiP15} consider an image to be a 2D sequence of image pixels, and uses a long short-term memory (LSTM) network to classify each pixel as background and foreground. 
In~\cite{binary_attention_icdar19}, Peng~\textit{et~al.} proposed a multi-resolutional attention model to learn the relationship between the text regions and background through convolutional conditional random field~\cite{krahenbuhl11a,teichmann19a}. 
%A U-Nets based network along with attention layer is designed for document image binarization in~\cite{binary_attention_icdar19}. 
%In order to capture feature at multiple scale, an input image is passed through this network at different scale. 
%Then these features are concatenated and passed through attention layer and softmax layer to obtain the final binary image. 
To bypass the need of large training datasets with ground truths, Kang~\textit{et~al.}~\cite{cascade_unet_binary_icdar19} employed modular U-Nets~\cite{ronneberger15a} pre-trained for specific tasks such as dilation, erosion, histogram equalization, etc. These U-Nets are cascaded using inter-module skip connections and the final network is fine-tuned for the document image binarization task. \\
% Kang~\textit{et~al.}~\cite{cascade_unet_binary_icdar19} proposes cascaded U-Nets architecture for the binarization task, where each U-Nets are pre-trained for specific tasks like dilation, erosion, histogram equalization, etc. These U-Nets are cascaded using inter-module skip connected to form the final network architecture. Finally, the cascaded network is fine-tuned for the binarization task. 

% \subsection{Document image enhancement}

\noindent\textbf{Document image enhancement}: 
In addition to working within the background foreground separation framework, existing works have developed noise-specific document image cleanup methods such as shadow removal. Bako~\textit{et~al.}~\cite{Bako16} assumes a constant background color generates a shadow map that matches local background colors to a global reference. Similar to Bako's method, local and background colors are estimated to remove shadow from document images in ~\cite{shadow_icip2019,shadow_icassp20}. Inspired by the topological surface filled by water, Jung~\textit{et~al.} proposed an illumination correction algorithm for document images in~\cite{Water-Filling_ACCV18}. A document image enhancement approach have been proposed by Krigler~\textit{et~al.} by representing the input image as 3D point cloud and adopting the visibility detection technique to detect the pixels to enhance~\cite{doc_enhance_pointcloud_cvpr18}. Recently, Lin~\textit{et~al.}~\cite{BEDSR-Net_CVPR20} proposed a deep architecture to estimate (i) the global background color of the document, and (ii) an attention map which computes the probability of a pixel belonging to the shadow-free background. An illumination correction and document rectification technique using patch based encoder-decoder network is proposed in~\cite{illu_cor2019}. 

Existing works have also explored deep networks for overall document enhancement rather than focusing on correcting specific document degradations. 
A skip-connected based deep convolutional auto-encoder is proposed in~\cite{Skip-Connected_ICPR18}. 
Instead of learning the transformation function  from input to output, this network learns the residual between input and output. 
This residual when subtracted from the input image results in a noise free enhanced image. 
An end to end document enhancement framework using conditional Generative Adversarial Networks (cGAN) is proposed in~\cite{DEGAN_TPAMI20}, where an U-Net based encoder-decoder architecture is used for the generator network. \\

\noindent\textbf{Document image cleanup for mobile and embedded applications}: Low-resource consuming models are desirable for mobile document image processing, e.g., in apps like Adobe Lens, CamScanner, Microsoft Office Lens, etc. However, existing CNN based methods~\cite{DEGAN_TPAMI20,Skip-Connected_ICPR18}, discussed above, propose deep architectures with huge number of parameters, making them unsuitable for memory and energy constrained devices. 
In this work, we propose a comparatively light-weight deep encoder-decoder based network for document image enhancement task. We employ the perceptual loss based transfer learning technique to compensate for generalization performance with a low network capacity. 
% obtain a favorable trade-off between resource/latency and accuracy. 
% The proposed network design and algorithm are discussed in the following section. 

% However, these networks are with huge numbers of parameters and hence they are not suitable for memory constrained devices. 
% To overcome this disadvantage, in this work we proposed a comparatively shallow encoder-decoder based network for document image enhancement task. 
%To learn the transformation function from input to output in a shallow network, the perceptual loss based transfer learning technique is utilized. Network design and loss function of our proposed method are discussed in the next section. \alertbyPJ{This is an important paragraph and we should strengthen it.}

\section{Proposed approach}
\label{sec:method}

As discussed, we propose a light-weight deep network, suitable for mobile document image cleanup applications. 
%Overall network design is shown in Fig.~\ref{fig:CNN_model}. 

% The network is designed in such a way so that the overall size of the network remain small.  
% The network parameters are learned by optimizing the perceptual loss function described in the following sections. 

\subsection{Network architecture}
\label{sub-sec:NA}
We design an encoder decoder based image to image translation network. 
%The image to image translation network we design for the document image enhancement task is an encoder decoder based model. 
The encoder part of the model consists of three convolution layers followed by five residual blocks. 
The residual blocks were first introduced in~\cite{RESNET_CVPR2016} for generic image processing tasks. 
We modify the residual blocks from the original design~\cite{RESNET_CVPR2016} to suite our network design. 
The decoder part of the model consists of five convolution layer along with skip connections from the encoder layers. 
This type of skip connection helps mitigate the vanishing gradient and the exploding gradient issues~\cite{RESNET_CVPR2016,Skip-Connected_ICPR18,DEGAN_TPAMI20}. 
Hence, the skip connections help to simplify the overall learning of the network. 
Each convolution layer is followed by batch normalization layer and ReLU6~\cite{relu6} activation layer. 

The kernel size of the convolution layer is $3 \times 3$ and strides for all the layers is set to $1$. 
The padding at each layer is set as {\it ``same''}, which helps to pad the input such that it is fully covered by the filter. 
Padding {\it ``same''} with stride $1$ helps to keep the spatial dimension of the convolution layer output same as its input. 
At the end of the network a sigmoid activation function is used to obtain a normalized output between $0$ and $1$. 
The output dimension of the last layer of the decoder is either one or three depending on the end task of the network. 
If the network is trained for the task of binarization or gray scale cleanup then the output dimension of the last layer is set to one. 
For color cleanup task the output dimension of the last layer is set to three.

\begin{figure}[t]
 \centering
 \fbox{\includegraphics[width=.9\textwidth]{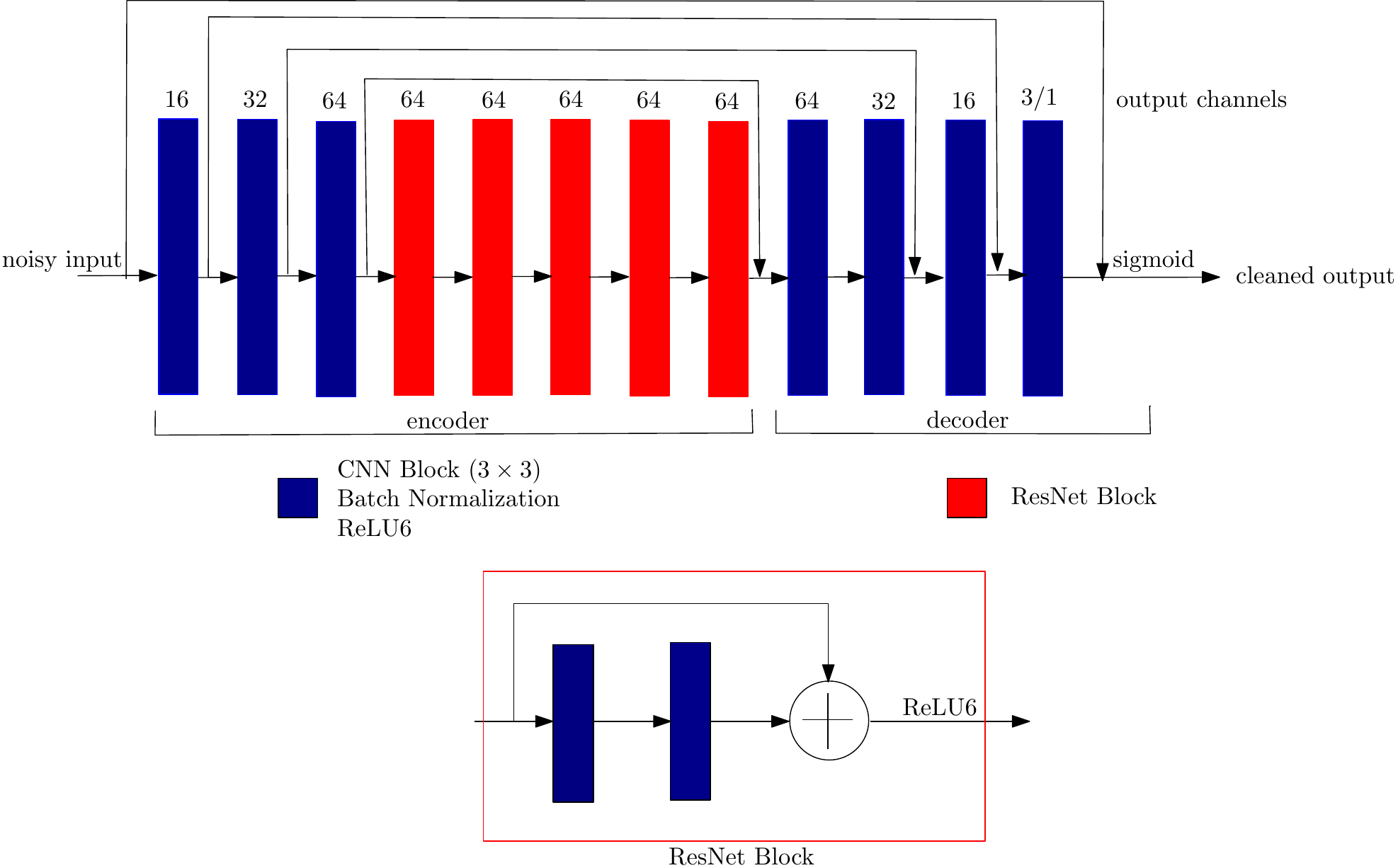}} 
 % LayoutTypes.pdf: 595x842 pixel, 72dpi, 20.99x29.70 cm, bb=0 0 595 842
 \caption{Proposed light-weight CNN architecture used for document image cleanup}
 \label{fig:CNN_model}
\end{figure}

We term our models as M-x, where x represents the value of the maximum width of the network. In our experiments, we have considered x to be $16$, $32$ and $64$. 
The M-64 model is shown in Fig.~\ref{fig:CNN_model}. 
In M-32 model's architecture, the output dimensions of the residual blocks are $32$ and the CNN blocks with output dimensions $64$ are removed. 
Similarly, in the case of M-16, the output dimensions of the residual blocks are set to 16 and the CNN blocks with output dimensions $32$ and $64$ are removed.

%The proposed network architecture is shown in Fig.~\ref{fig:CNN_model}. 

\subsection{Loss function}
\label{sub-sec:LF}

The network is optimized by minimizing the loss function $L$ computed using Eq.~\ref{eq:loss}: 
\begin{equation}
L(I_t,I_g) = \lambda_1 \ell_1(I_t,I_g)  + \lambda_2 \ell_2(I_t,I_g)  + \lambda_3 \ell_3(I_t,I_g). 
\label{eq:loss}
\end{equation}
Here, $\ell_1(I_t,I_g)=\| I_t - I_g \|_1$ is the $1$-norm loss between the translated image $I_t$ and the ground truth image $I_g$ in $YC_bC_r$ color space for color image cleanup. 
For gray scale cleanup, $\ell_1(I_t,I_g)=\| I_t - I_g \|_1$, refer to the $1$-norm loss between the two images in gray scale. 
%The loss functions $\ell_2(I_t,I_g)$ and $\ell_3(I_t,I_g)$ represent perceptual loss between $I_t$ and $I_g$~\cite{Johnson2016Perceptual}. 
In addition to pixel-level loss function $\ell_1(I_t,I_g)$, we also employ the perceptual loss functions $\ell_2(I_t,I_g)$ and $\ell_3(I_t,I_g)$ in Eq.~\ref{eq:loss}. 

Perceptual loss functions~\cite{Johnson2016Perceptual,rad19a,zhao17a} compute the difference between images $I_t$ and $I_g$ at high level feature representations extracted from a pre-trained CNNs such as those trained on ImageNet image classification task. They are more robust in computing distance between images than pixel-level loss functions. 
In the context of developing light-weight document image cleanup models, perceptual loss functions serve an additional role of enabling transfer learning. The perceptual loss functions in Eq.~\ref{eq:loss} helps to transfer the semantic knowledge already learned by the pre-trained CNN network to our smaller network. 
% Instead of training a high capacity CNN model for such image translation tasks, the perceptual loss function helps to transfer the semantic knowledge already learned by the pre-trained CNN deeper network to our smaller network. 
% Since there is no definite single outcome of image translation in case of the cleanup task, its requires semantic reasoning between the translated and ground truth image. 
% In theory, a high capacity neural network trained for such image translation task may solve the problem. 
% However, we need not need to learn from scratch. 
% The perceptual loss functions help to transfer the semantic knowledge already learned by a deeper network to our smaller network. 

The perceptual loss has two components~\cite{Johnson2016Perceptual}: feature reconstruction loss $\ell_2(I_t,I_g) $ and style loss $\ell_3(I_t,I_g)$. 
Feature reconstruction loss encourages the transformed image to be similar to ground truth image at high level feature representation as computed by a pre-trained network $\wp$. Let  $\wp_j(I)$ be the activations of the $j^{th}$ layer of the pre-trained network $\wp$.
Then, Eq~\ref{eq:fcl} represents the feature reconstruction loss:  
%can be represented using Eq~\ref{eq:fcl}, 
\begin{equation}
\ell_2(I_t,I_g) = \frac{1}{H_jW_jC_j}\|\wp_j(I_t) - \wp_j(I_g)\|_1,
\label{eq:fcl}
\end{equation}
where the shape of $\wp_j(I)$ is $H_j \times W_j \times C_j$. 
The feature reconstruction loss penalizes the transformed image when it deviate from the content of the ground truth image. 
Additionally, we should also penalize the transformed image if it deviate from the ground truth image in terms of common feature, texture, etc. 
To achieve this style loss is incorporated as proposed in~\cite{Johnson2016Perceptual}. 
The style loss is represented in Eq.~\ref{eq:sl} as follows:
\begin{equation}
\ell_3(I_t,I_g) = \sum_{\forall j \in J } \| G^{\wp}_{j}(I_t) -  G^{\wp}_{j}(I_t) \|,
\label{eq:sl}
\end{equation}
where $\wp$ represent pre-trained CNN network, $J$ represent set of layers of $\wp$ used to compute style loss, and $G^{\wp}_{j}(I)$ represent a Gram matrix containing second-order feature covariances. 
Let $\wp_j(I)$ be the activation of the $j^{th}$ layer of the pre-trained network $\wp$, where the shape of $\wp_j(I)$ is $H_j \times W_j \times C_j$. Then, the shape of the Gram matrix $G^{\wp}_{j}(I)$ is $C_j \times C_j$ and each element of $G^{\wp}_{j}(I)$ is computed according to Eq~\ref{eq:gram} as follows: 
\begin{equation}
G^{\wp}_{j}(I)_{c,c'} = \frac{1}{H_jW_jC_j}\sum_{h=1}^{H_j}\sum_{w=1}^{W_j} \wp_j(I)_{h,w,c} \wp_j(I)_{h,w,c'}. 
\label{eq:gram}
\end{equation}
In our work, we use the $VGG19$ network~\cite{simonyan2014deep} trained on the ImageNet classification task~\cite{imagenet_cvpr09} as our pre-trained network $\wp$ for the perceptual loss. 
Here, feature reconstruction loss is computed at layer conv1-2 and style reconstruction loss is computed at layers conv1-1, conv2-1, conv3-1, conv4-1, and conv5-1.

\section{Experimental results and discussion}
\label{sec:result}

We evaluate the generalization performance of the proposed models on binarization, gray scale, and color cleanup tasks. 

%In this section, we present our experimental results. 
%To show the effectiveness of the proposed methods, we have conducted our experimentation along three dimensions namely, binarization, gray and color scale cleanup.  

\begin{table}[t]
\begin{center}
\setlength{\tabcolsep}{3pt}
\begin{tabular}{@{}lS[table-format=3.1]S[table-format=2.2]S[table-format=6.0]S[table-format=1.2]S[table-format=1.2]@{}}
\toprule
\multirow{2}{*}{Model} &   {Mult-Adds} & {Parameters} & {Size} & {Inference time} & {Load time} \\ %\hline
    &       {(in billions)} & {(in millions)} & {(in KB)} & {(in seconds)} & {(in seconds)} \\ 
\midrule    
DE-GAN~\cite{DEGAN_TPAMI20} & 46.1 & 31.00 & 121215 & 0.36 & 9.96 \\ %\hline
SkipNetModel~\cite{Skip-Connected_ICPR18} & 106.8 & 1.64 & 6380 & 0.34 & 3.62 \\ %\hline
M-64 (proposed)  & 15.1 & 0.46 & 1779 & 0.24 & 3.25 \\ %\hline % 0.456
M-32 (proposed)  & 6.7 & 0.11 & 445 & 0.07 & 2.89 \\ %\hline % 0.113
M-16 (proposed)  & 1.7 & 0.03 & 111 & 0.02 & 2.68 \\ % 0.027
\bottomrule
\end{tabular}
\caption{No. of parameters, product-sum operations, and other statistics of various models. Average per-patch inference time is reported. As an example, a $2560\times 2560$ image has 100 patches. }
\label{tab:ops_params}
\end{center}
\end{table}

% \begin{table}[t]
% \begin{center}
% \setlength{\tabcolsep}{4pt}
% \begin{tabular}{lS[table-format=5.0]S[table-format=2.2]}
% \toprule
% Model &   {Billion Mult-Adds} & {Million parameters} \\ %\hline
% \midrule    
% DE-GAN~\cite{DEGAN_TPAMI20} & 46.1 & 31.00   \\ %\hline
% SkipNetModel~\cite{Skip-Connected_ICPR18} & 106.8 & 1.64  \\ %\hline
% M-64 (proposed)  & 15.1 & 0.46 \\ %\hline % 0.456
% M-32 (proposed)  & 6.7 & 0.11  \\ %\hline % 0.113
% M-16 (proposed)  & 1.7 & 0.03  \\ % 0.027
% \bottomrule
% \end{tabular}
% \caption{No. of parameters and product-sum operations for each model}
% \label{tab:ops_params}
% \end{center}
% \end{table}

\noindent\textbf{Experimental setup}: In our experiment, the input of the network is set as $256 \times 256$. 
The input to the network is a $3$ channel RGB image, whereas the output dimension of the network is set as $1$ or $3$ depending on the downstream task. 
If the downstream task is to obtain an image in gray scale or a binary image, then the output dimension is set as $1$. 
The output dimension is set as $3$ for color cleanup task. 

To handle different type of noise at various resolution, the training images are scaled at scale $0.7$, $1.0$, and $1.4$. 
Further at each scale, the training images are divided into overlapping blocks of $256 \times 256$. 
During training, a few random patches from the training images are also used for data augmentation using random brightness-contrast, jpeg noise, ISO noise, and various types of blur~\cite{albumination}. 
Randomly selected $80\%$ of the training patches is used to train the model while the remaining $20\%$ is kept for validation. 
%The set of training patches is further randomly divided into $80\%$ training and $20\%$ validation. 
The model with best validation performance is saved as the final model. 
The network is optimized using Adam algorithm~\cite{adam} with default parameter settings. 
The parameters $\lambda_1$, $\lambda_2$, and $\lambda_3$ of Eq~\ref{eq:loss} is set to $1e1$, $1e-1$, and $1e1$ respectively. 
During inference, an input image is divided into overlapping blocks of $256 \times 256$. Each patch is inferred using the trained model. Finally, all the patches are merged to obtain the final result. We use simple averaging for the overlapping pixels of the patches. \\
%To show the effectiveness of the proposed method, we have conducted our experimentation on three tasks namely, binarization, gray scale cleanup, and color cleanup.  

\noindent\textbf{Compared algorithms}: 
%We compare the proposed model with existing works 
%Existing works mainly focus on deep 
We compare the proposed models with recently proposed deep CNN based document image cleanup models: SkipNetModel~\cite{Skip-Connected_ICPR18} and DE-GAN~\cite{DEGAN_TPAMI20}. 
%To the best of our knowledge, there are no DNN based method which is light-weight in nature designed for document cleanup task. 
% Hence, the proposed model is compared with the models whose inference networks are deeper in nature but similar in network design i.e., encoder-decoder based models. 
% The proposed model is compared with the models proposed in~\cite{Skip-Connected_ICPR18} and~\cite{DEGAN_TPAMI20}. 
% For the evaluation purpose, we have implemented and trained the model~\cite{Skip-Connected_ICPR18}. 
% We refer to the models proposed in~\cite{Skip-Connected_ICPR18} and~\cite{DEGAN_TPAMI20}  as SkipNetModel and DE-GAN respectively. 
Table~\ref{tab:ops_params} presents a comparative analysis of our proposed models with SkipNetModel and DE-GAN in terms of: (i) number of multiplication and addition operations (Mult-Adds) associated with the model~\cite{howard17a}, (ii) number of parameters, (iii) actual size on device, (iv) model load time, and (v) model inference time. A comparison with respect to these parameters is essential if the applicability of any model for memory and energy constrained devices is to determined. We implemented the models using TensorFlow Lite (\url{https://www.tensorflow.org/lite}) on an Android device with Qualcomm SM8150 Snapdragon 855 chipset and 6GB RAM size. 
On the device, we observe that our models are 65-1090 and 3-55 times lighter in size than DE-GAN and SkipNetModel, respectively.
%We observe that on-device size of our proposed models are 65-1090 and 3-55 times better than DE-GAN and SkipNetModel, respectively. 
Similarly, our models has lesser product-sum operations and prediction time during the inference stage, making them suitable to mobile and embedded applications. 
%Smaller size and lesser product-sum operations enable swift prediction for models during the inference stage, making them suitable to mobile and embedded applications. 
%Similarly, our models involve much less product-sum operations, making them suitable to mobile and embedded applications. 
% their suitability to mobile and embedded applications. 

%In our experimentation, we have focused on comparing our proposed model with  state-of-the-art with regard to the following aspects - (i) number of parameters, and (ii) number of multiplication and addition operations (Mult-Adds) associated with the model~\cite{howard17a}. A comparison with respect to these parameters is essential if the applicability of any model for memory and energy constrained devices is to determined. Table~\ref{tab:ops_params} presents a comparative analysis of our proposed models with aforementioned models with respect to the parameters discussed above.  

\begin{table}[t]
\begin{center}
\setlength{\tabcolsep}{4pt}
\begin{tabular}{lccccc}
\toprule
Model & F-measure & $F_{ps}$ & PSNR & DRD \\ %\hline
\midrule
Otsu~\cite{otsu} & $83.9$ & $86.5$ & $16.6$ & $11.0$ \\ %\hline
Sauvola~\textit{et.al.}~\cite{Sauvola00adaptivedocument} & $85.0$ & $89.8$ & $16.9$ & $7.6$ \\ %\hline \hline
Tensmeyer~\textit{et.al.}~\cite{Tensmeyer_Binary_ICDAR17} & $93.1$ & $96.8$ & $20.7$ & $2.2$ \\ %\hline
Vo~\textit{et.al.}~\cite{Vo_bin_PR18} & $94.4$ & $96.0$ & $21.4$ & $1.8$ \\ %\hline
DE-GAN~\cite{DEGAN_TPAMI20} & $99.5$ & $99.7$ & $24.9$ & $1.1$ \\ %\hline
SkipNetModel~\cite{Skip-Connected_ICPR18} & $95.3$ & $96.6$ & $22.8$ & $1.5$ \\ %\hline
M-64 (proposed)  & $94.1$ & $95.7$ & $21.7$ & $2.1$ \\ %\hline
M-32 (proposed) & $92.3$ & $93.3$ & $20.4$ & $2.5$ \\ %\hline
M-16 (proposed) & $90.4$ & $91.6$ & $19.9$ & $3.1$ \\ %\hline
\bottomrule
\end{tabular}
\end{center}
\caption{Results on DIBCO13~\cite{DIBCO13}}
\label{tab:dibco13}
\end{table}
\begin{figure}[b]\center
\begin{tabular}{@{}c@{\ }c@{\ }c@{\ }c@{}}
\fbox{\includegraphics[width=.22\textwidth]{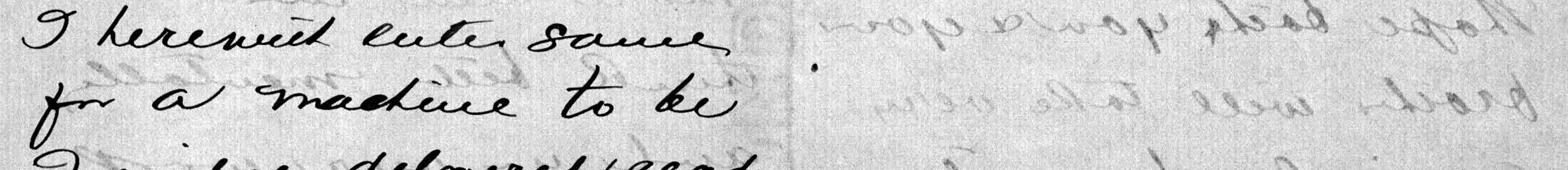}}&
\fbox{\includegraphics[width=.22\textwidth]{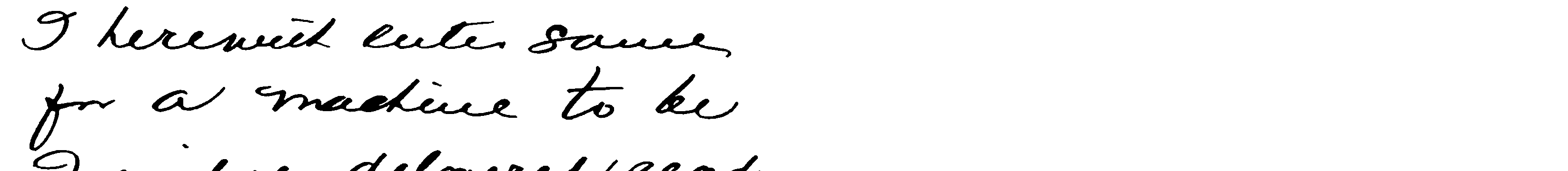}}&
\fbox{\includegraphics[width=.22\textwidth]{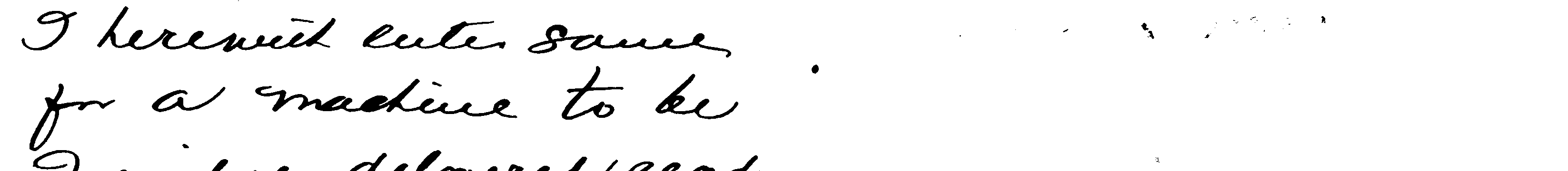}}&
\fbox{\includegraphics[width=.22\textwidth]{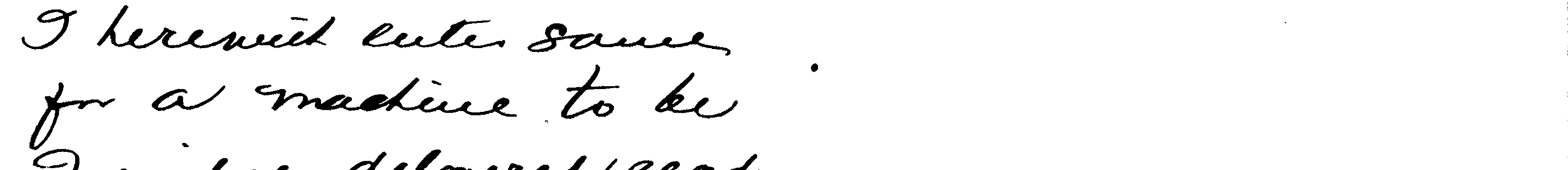}}\\
%\multicolumn{3}{c}{(a)} \\
%(a)&(b)&(c)
\fbox{\includegraphics[width=.22\textwidth]{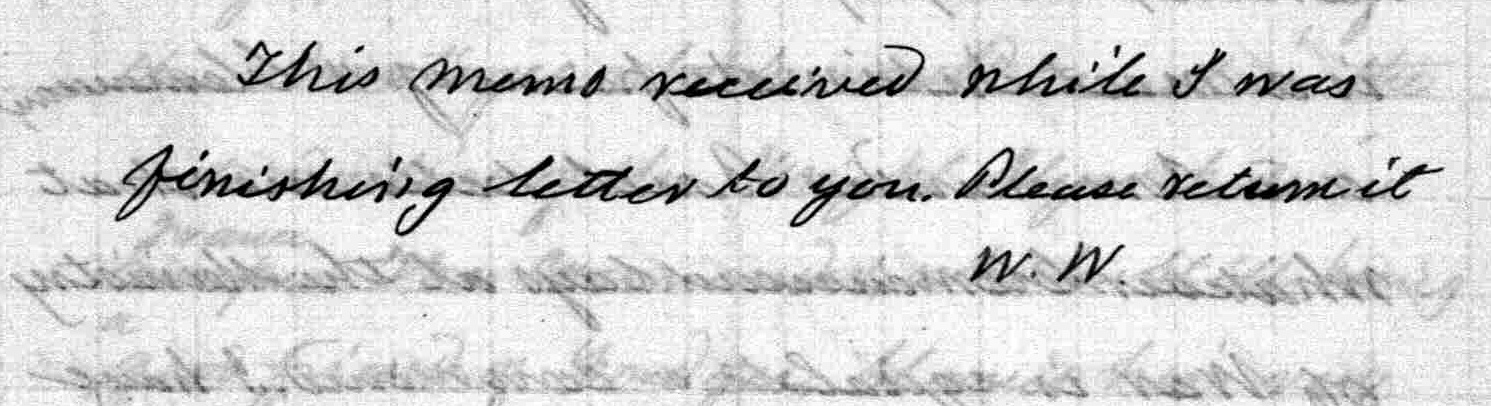}}&
\fbox{\includegraphics[width=.22\textwidth]{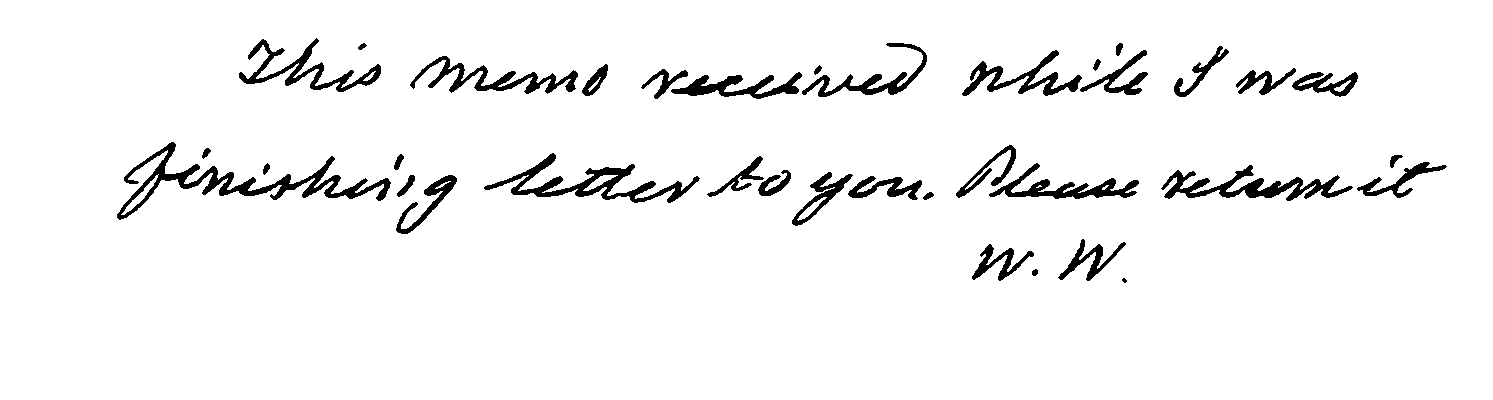}}&
\fbox{\includegraphics[width=.22\textwidth]{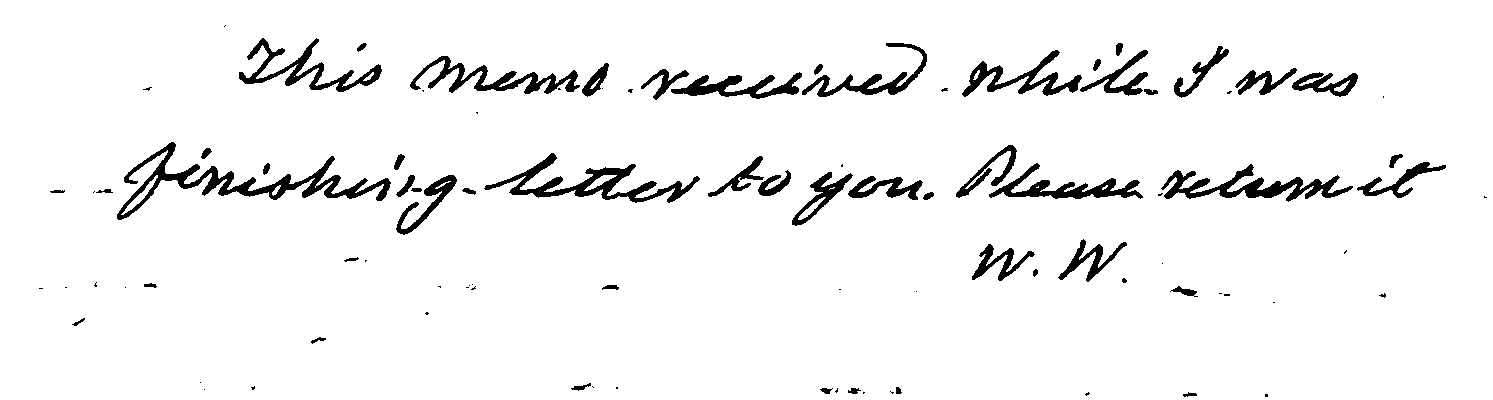}}&
\fbox{\includegraphics[width=.22\textwidth]{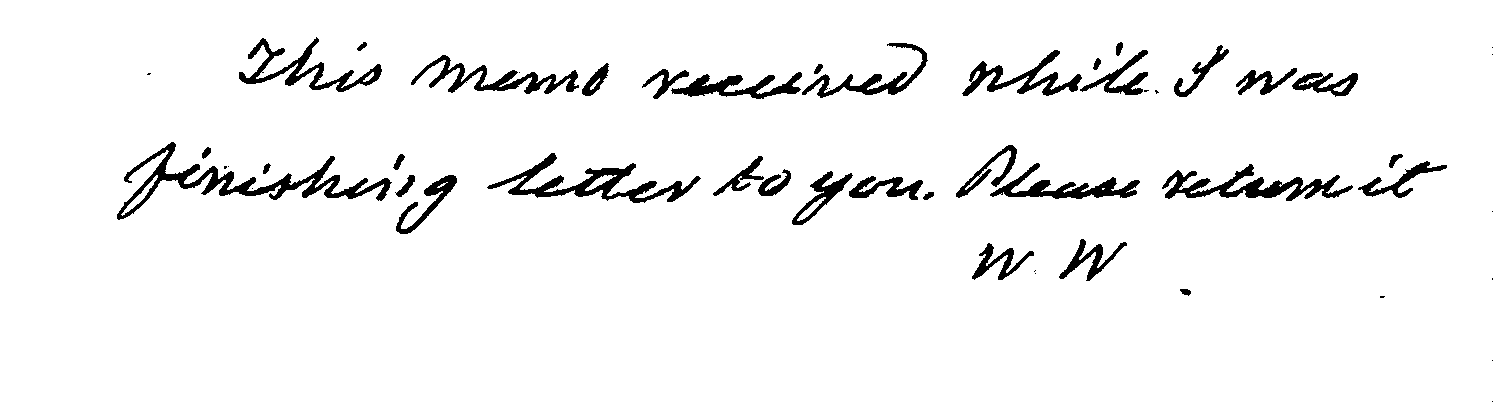}}\\

\fbox{\includegraphics[width=.22\textwidth]{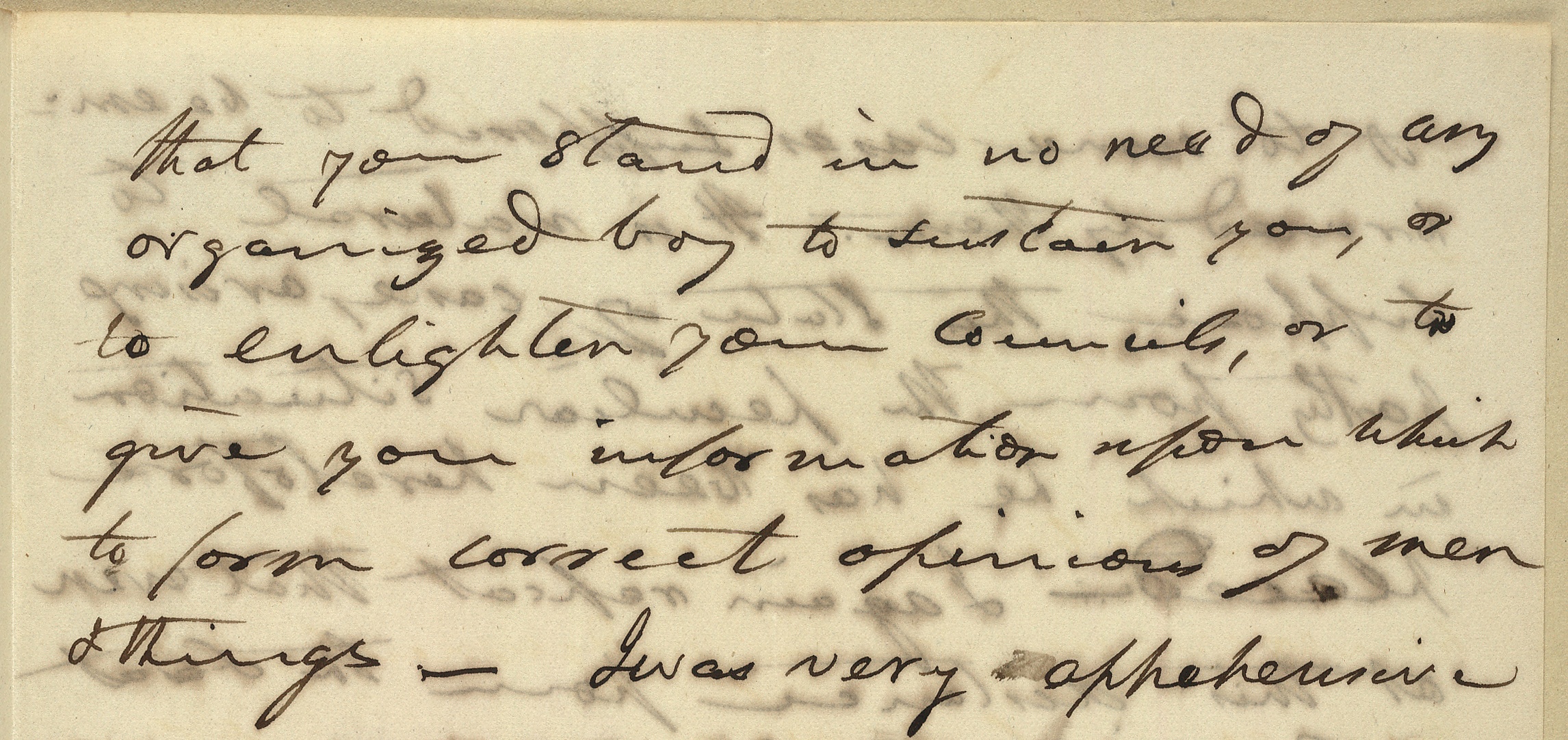}}&
\fbox{\includegraphics[width=.22\textwidth]{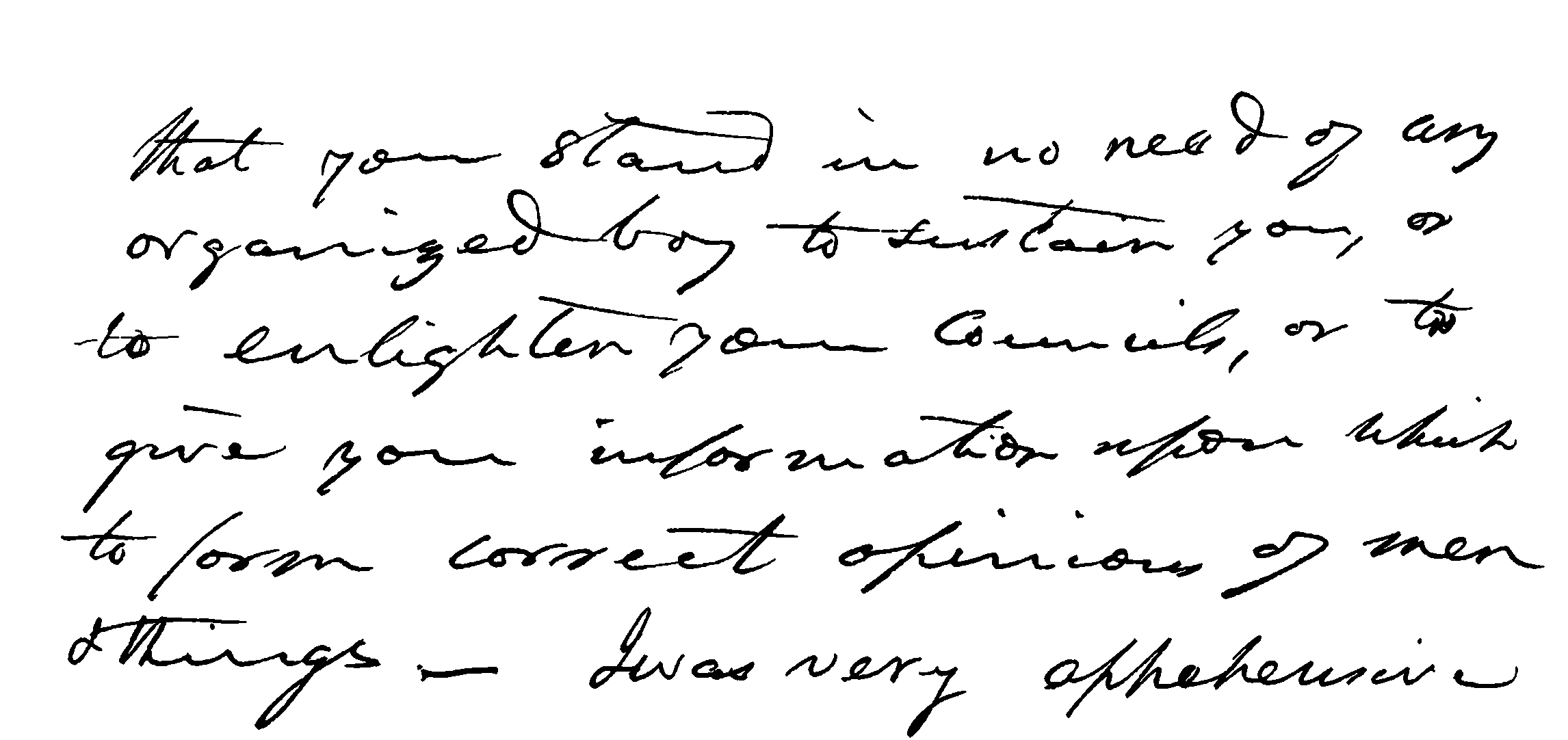}}&
\fbox{\includegraphics[width=.22\textwidth]{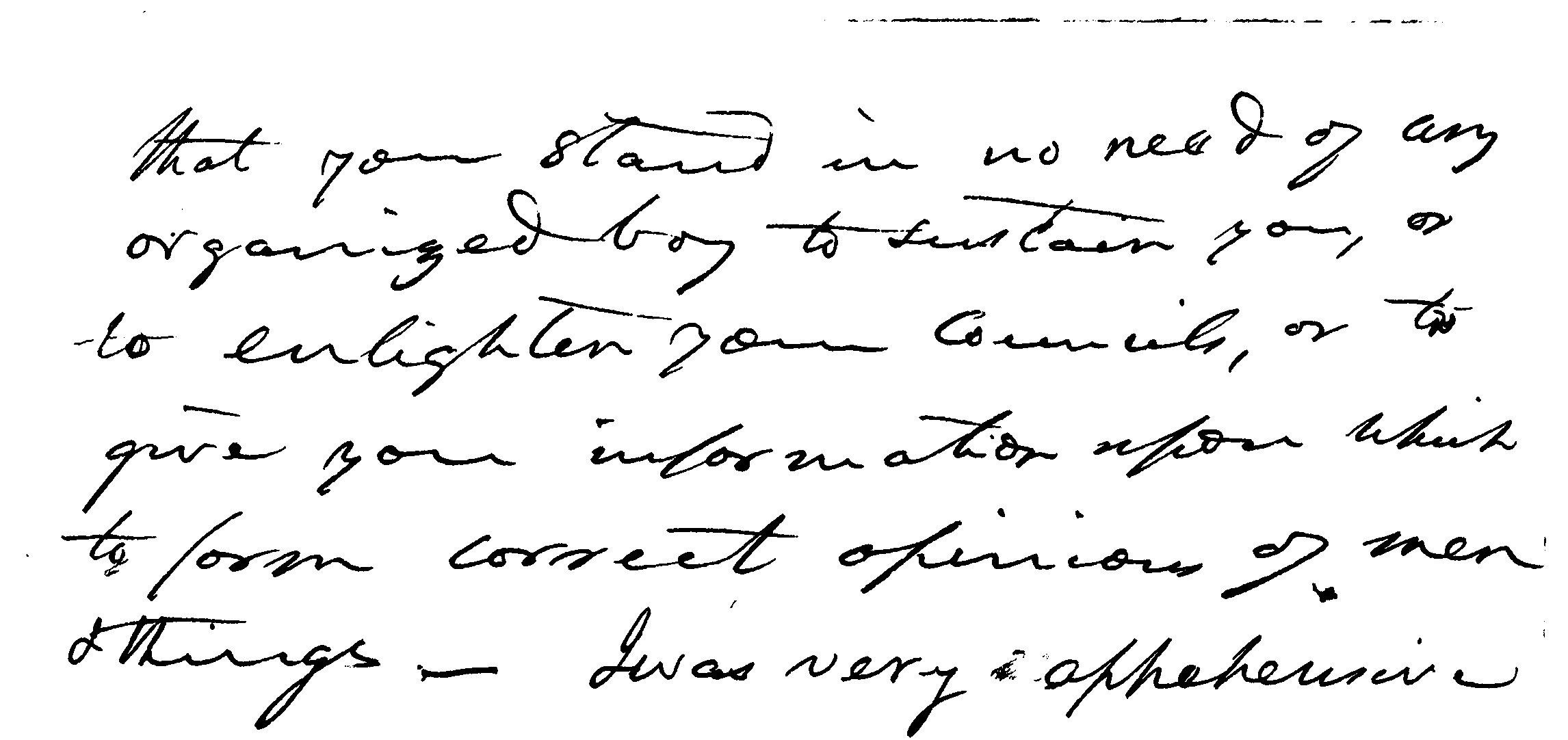}}&
\fbox{\includegraphics[width=.22\textwidth]{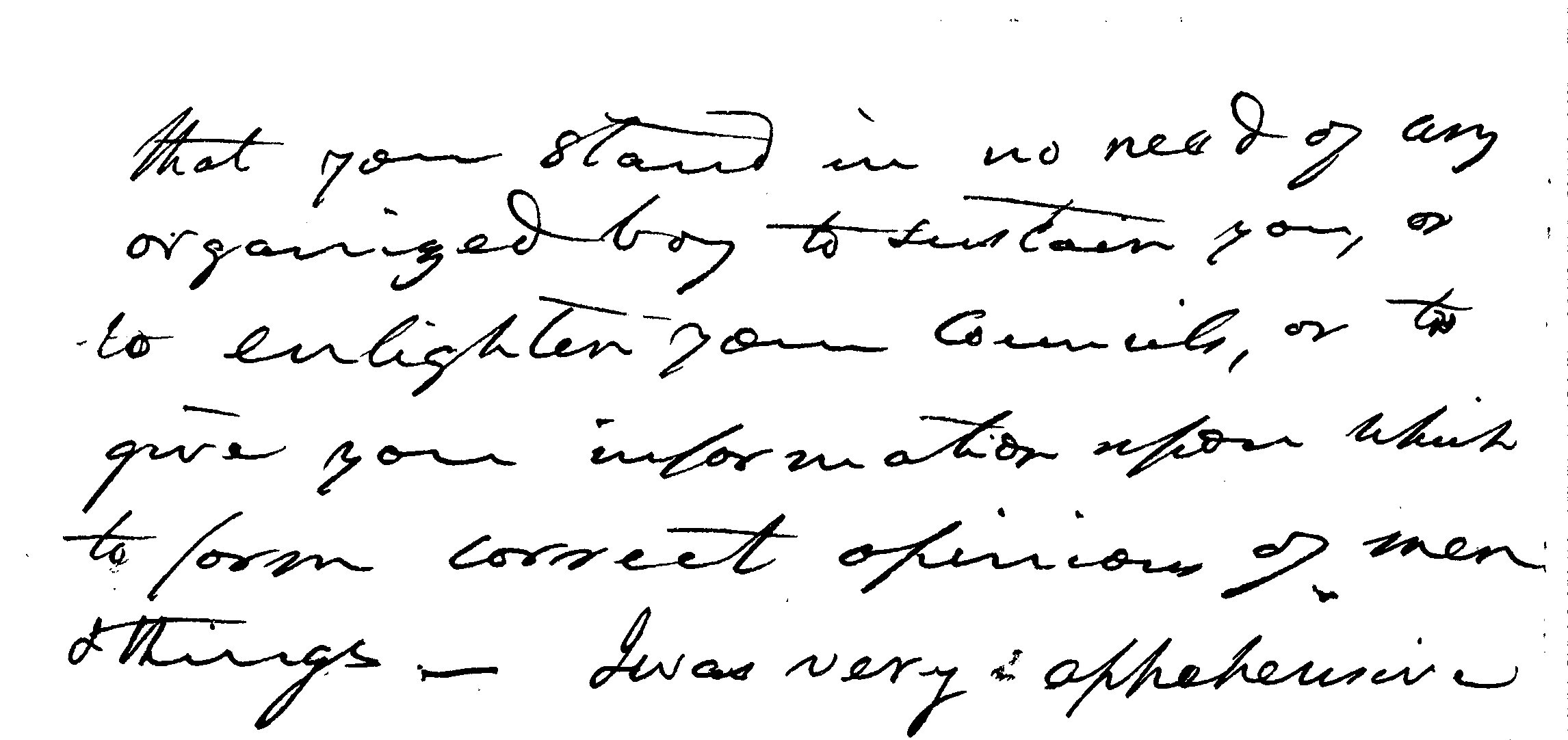}}\\
input & ground truth & SkipNetModel~\cite{Skip-Connected_ICPR18}  & M-32\\
\end{tabular}
\caption{Typical examples of DIBCO13~\cite{DIBCO13}. }
\label{fig:dibco13_example}
\end{figure}

% \begin{table}
% \begin{center}
% \begin{tabular}{|c|c|c|c|c|c|}
% \hline
% Model & F-measure & $F_{ps}$ & PSNR & DRD \\ \hline
% Otsu~\cite{otsu} & $83.9$ & $86.5$ & $16.6$ & $11.0$ \\ \hline
% Sauvola~\textit{et.al.}~\cite{Sauvola00adaptivedocument} & $85.0$ & $89.8$ & $16.9$ & $7.6$ \\ \hline \hline
% Tensmeyer~\textit{et.al.}~\cite{Tensmeyer_Binary_ICDAR17} & $93.1$ & $96.8$ & $20.7$ & $2.2$ \\ \hline
% Vo~\textit{et.al.}~\cite{Vo_bin_PR18} & $94.4$ & $96.0$ & $21.4$ & $1.8$ \\ \hline
% DE-GAN~\cite{DEGAN_TPAMI20} & $99.5$ & $99.7$ & $24.9$ & $1.1$ \\ \hline
% SkipNetModel~\cite{Skip-Connected_ICPR18} & $95.3$ & $96.6$ & $22.8$ & $1.5$ \\ \hline
% M-64  & $94.1$ & $95.7$ & $21.7$ & $2.1$ \\ \hline
% M-32  & $92.3$ & $93.3$ & $20.4$ & $2.5$ \\ \hline
% M-16  & $90.4$ & $91.6$ & $19.9$ & $3.1$ \\ \hline
% \end{tabular}
% \end{center}
% \caption{Results on DIBCO13~\cite{DIBCO13}}
% \label{tab:dibco13}
% \end{table}

\subsection{Binarization}

\begin{figure}[t]\center

\begin{tabular}{@{}c@{\ }c@{\ }c@{\ }c@{}}
\fbox{\includegraphics[width=.22\textwidth]{2_in.jpeg}}&
\fbox{\includegraphics[width=.22\textwidth]{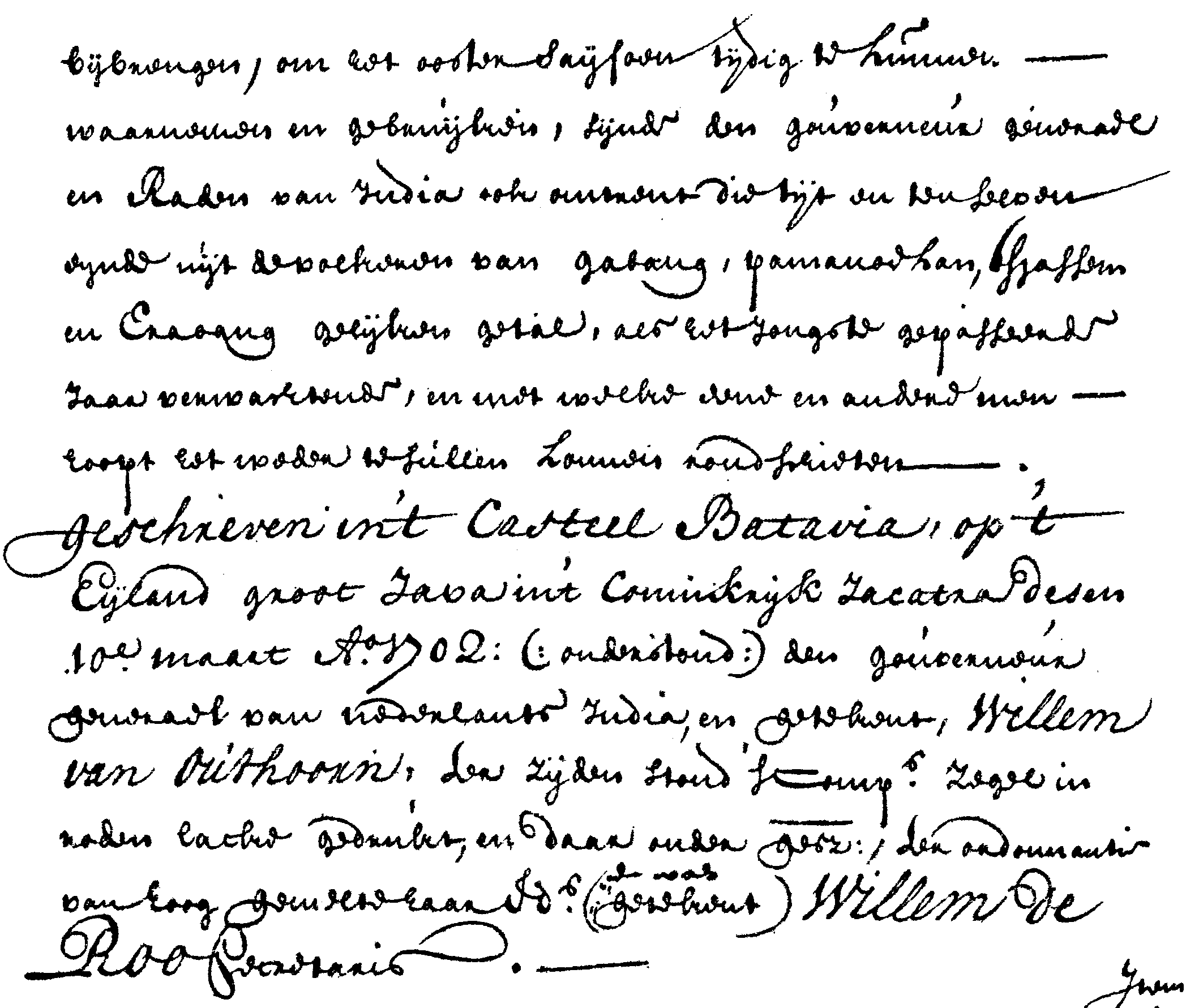}}&
\fbox{\includegraphics[width=.22\textwidth]{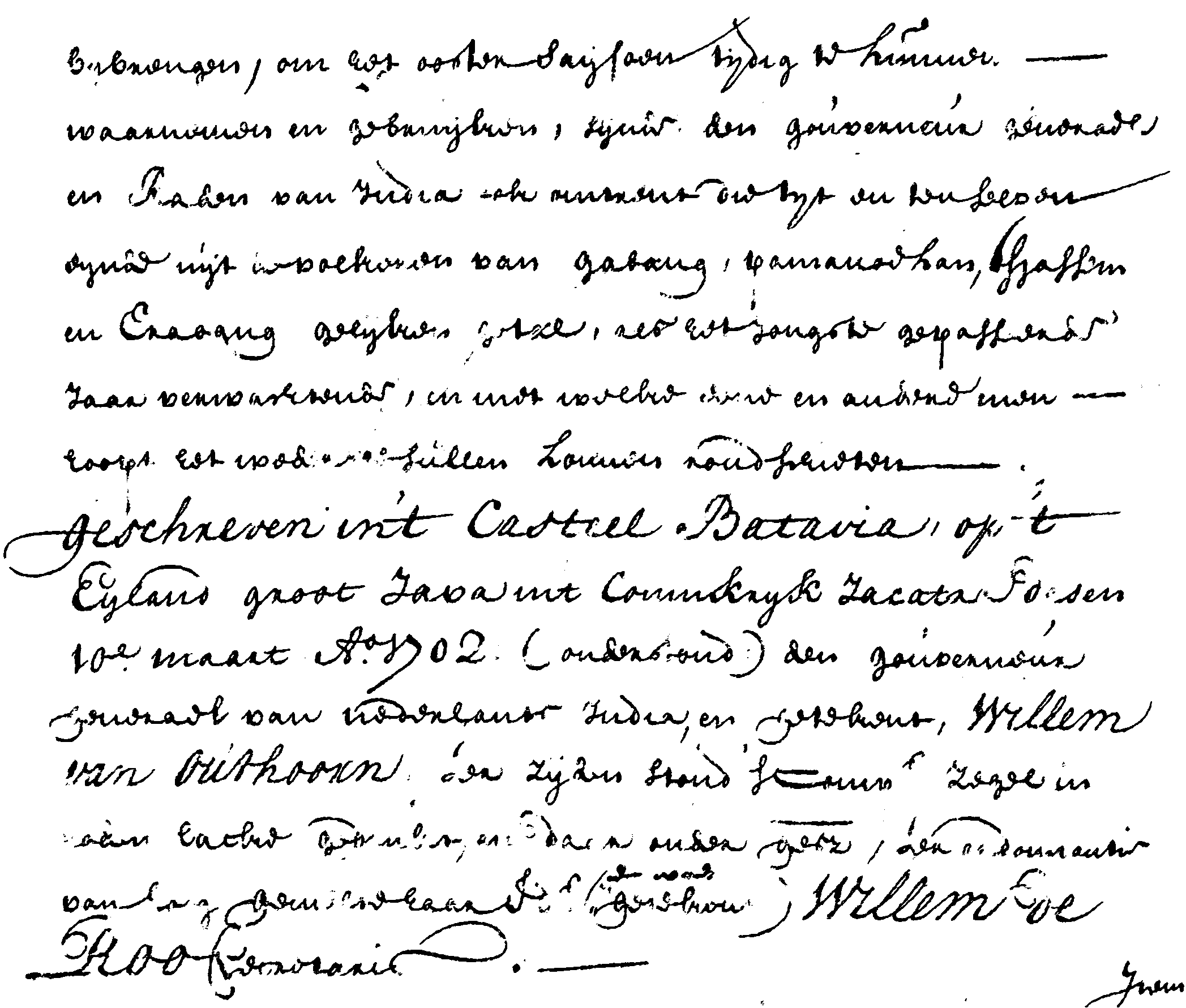}}&
\fbox{\includegraphics[width=.22\textwidth]{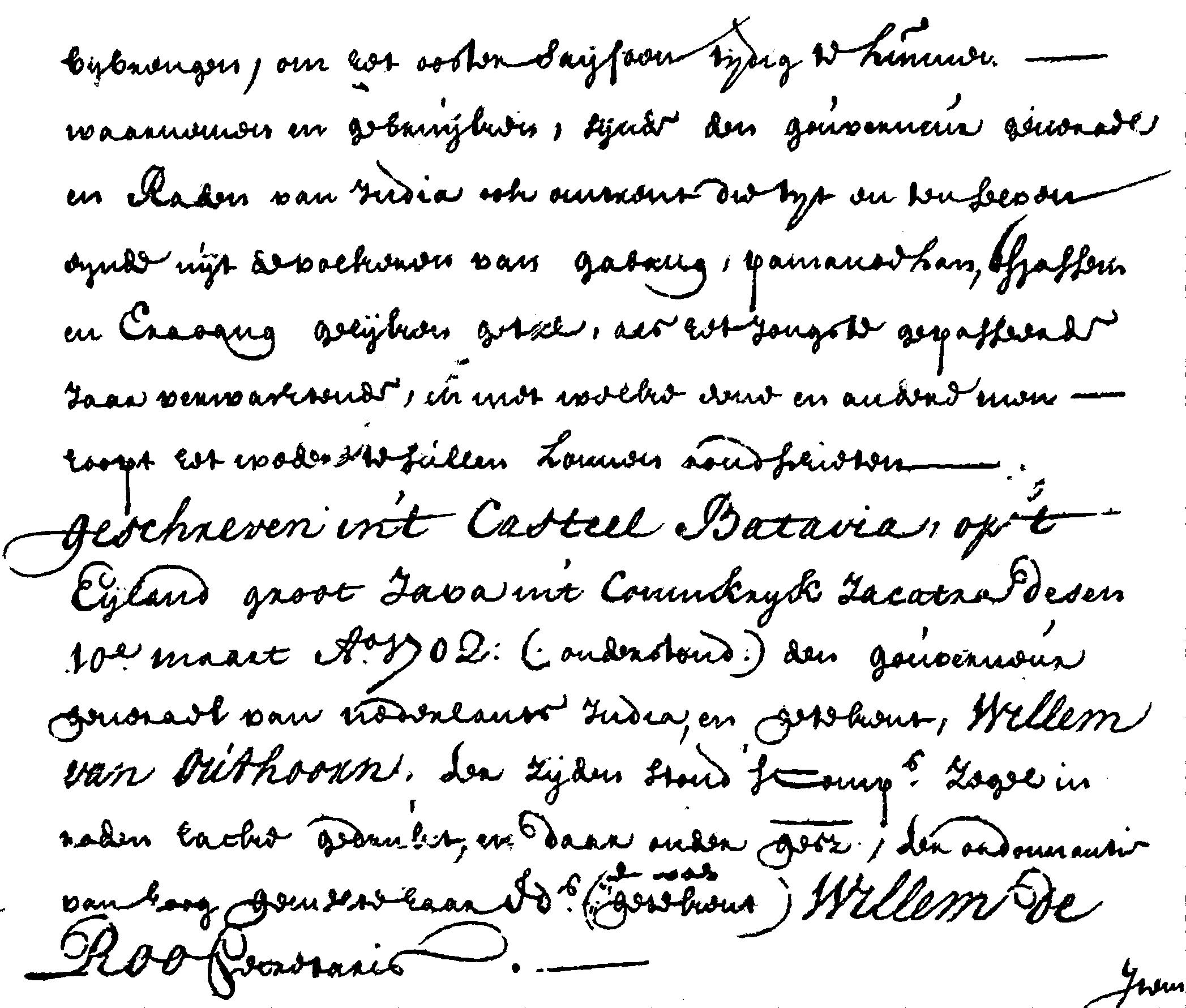}}\\
%\multicolumn{3}{c}{(a)} \\
%(a)&(b)&(c)
\fbox{\includegraphics[width=.22\textwidth]{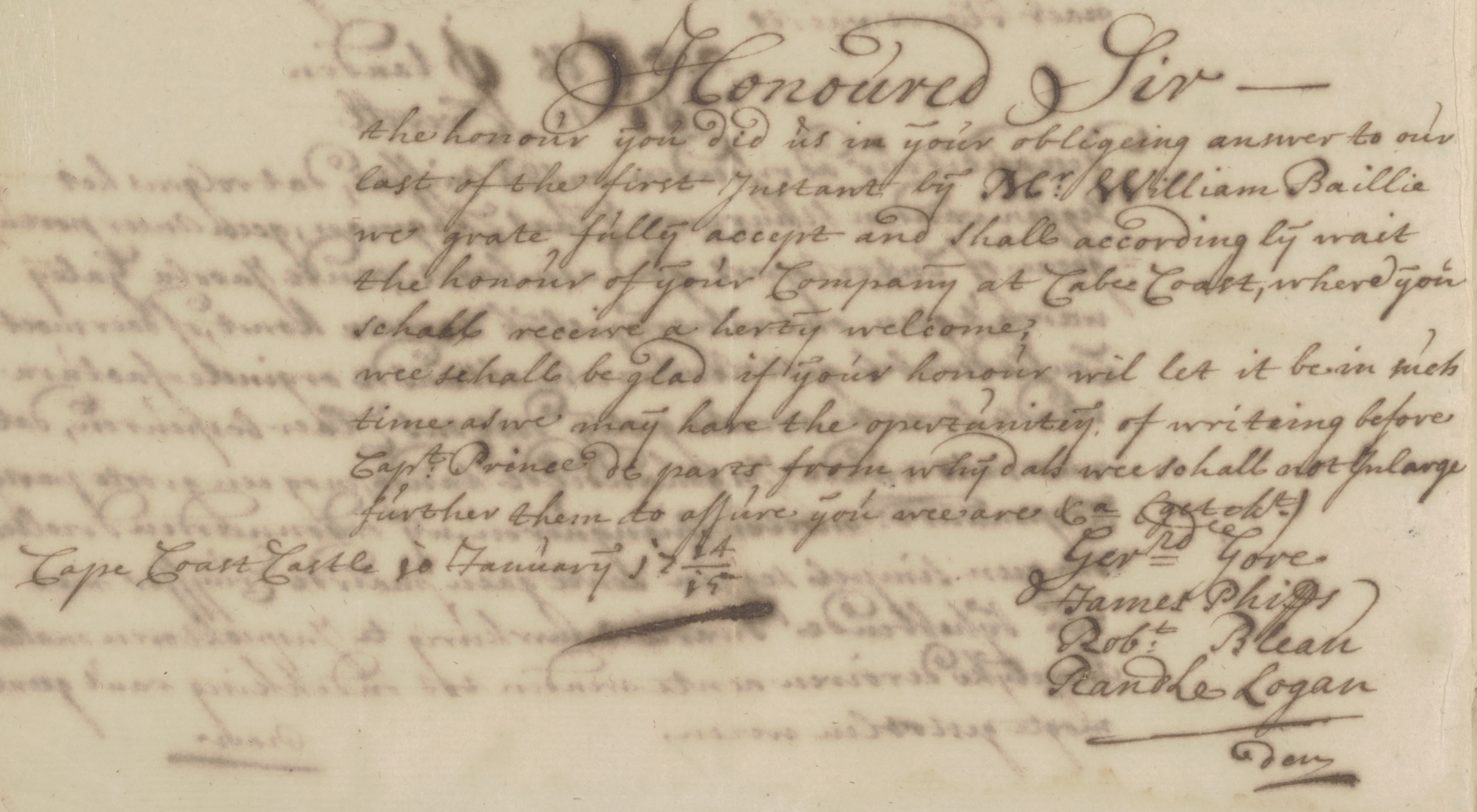}}&
\fbox{\includegraphics[width=.22\textwidth]{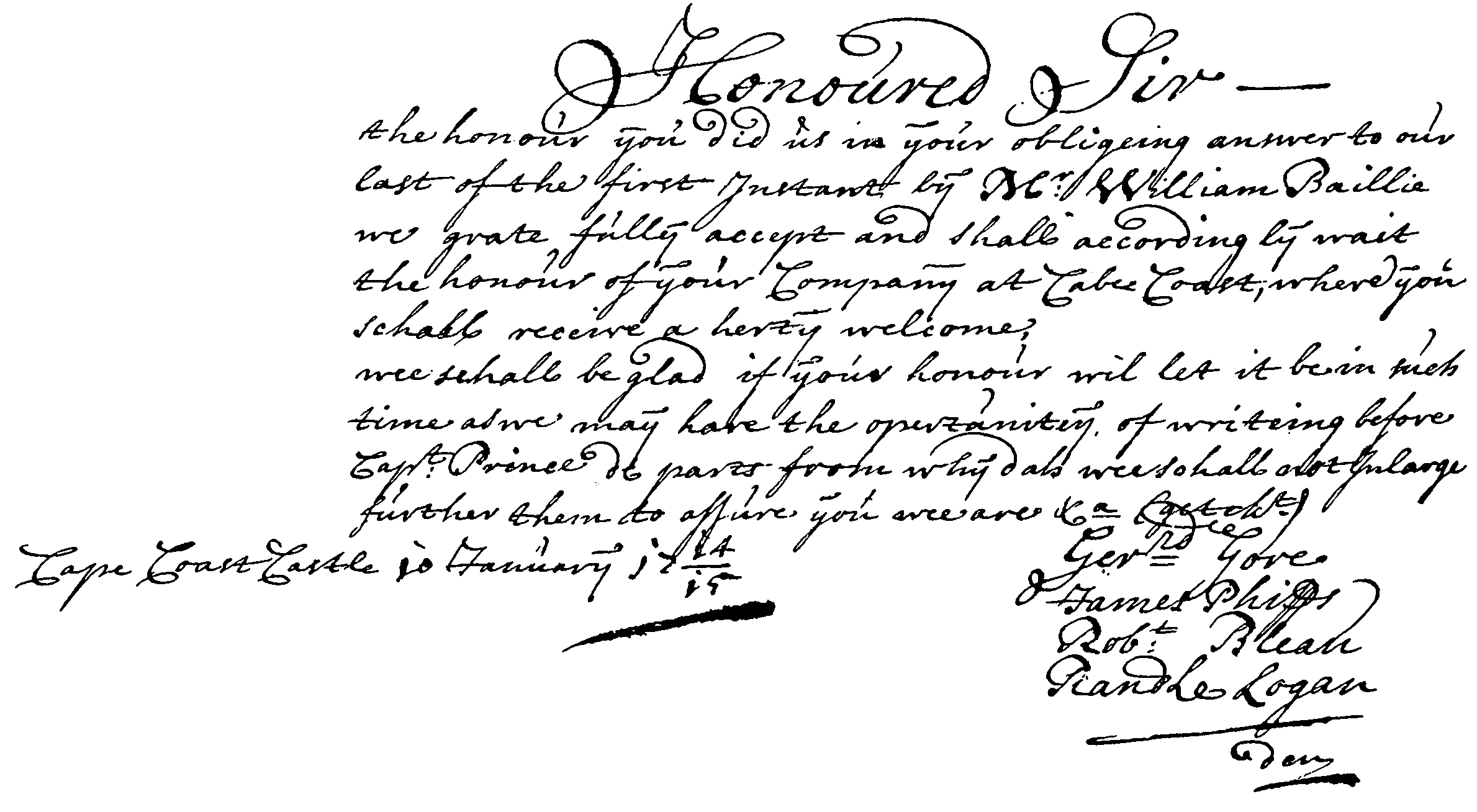}}&
\fbox{\includegraphics[width=.22\textwidth]{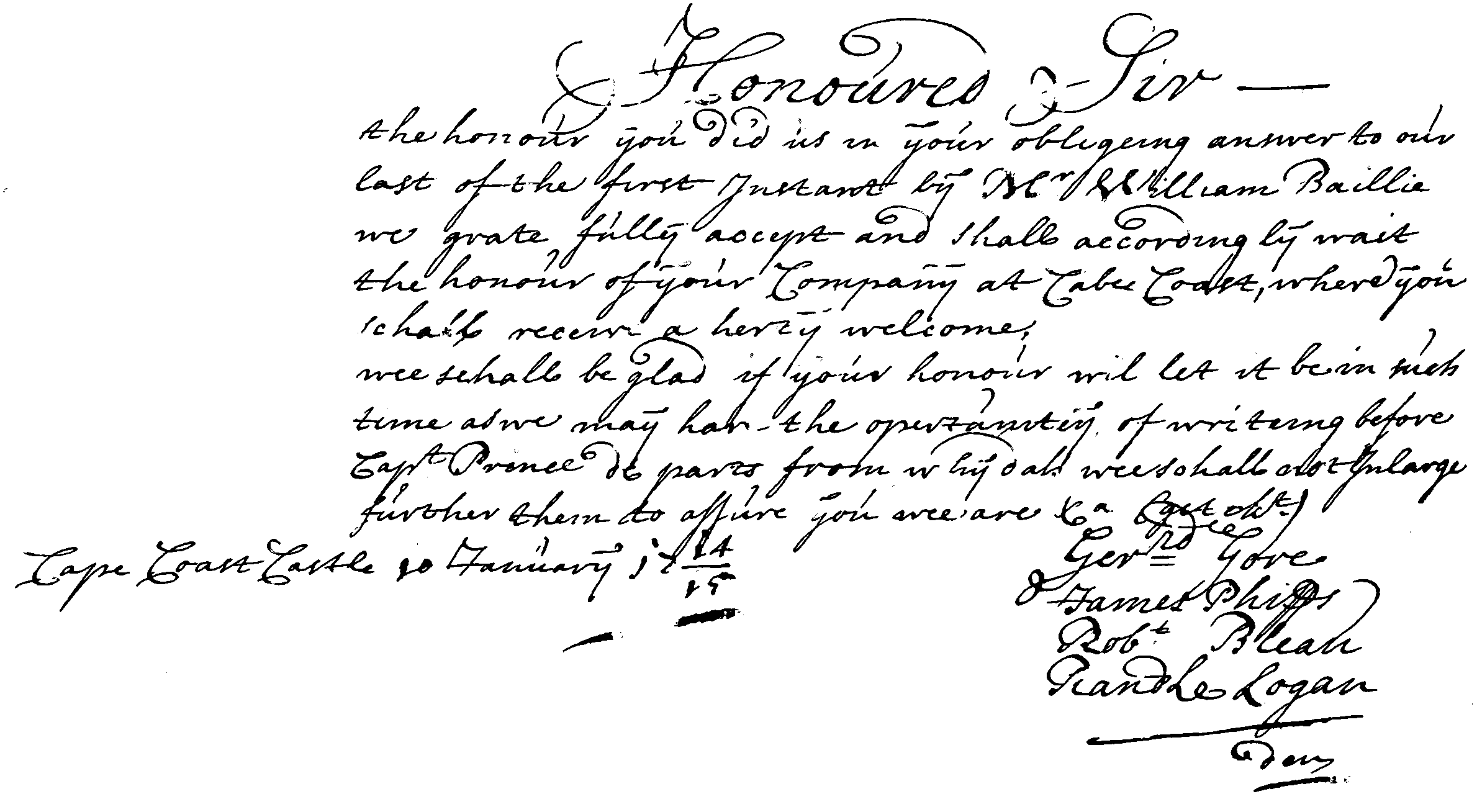}}&
\fbox{\includegraphics[width=.22\textwidth]{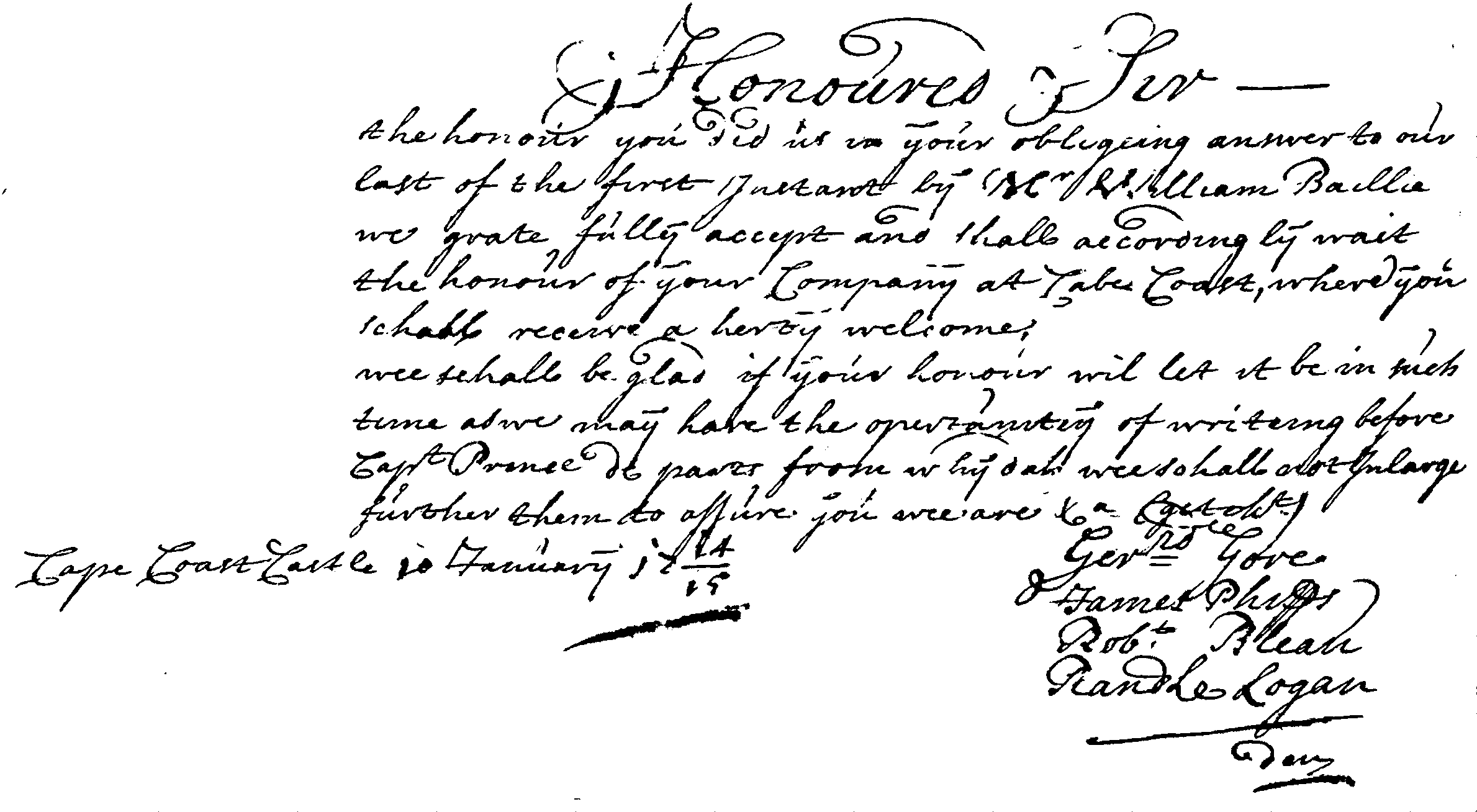}}\\

\fbox{\includegraphics[width=.22\textwidth]{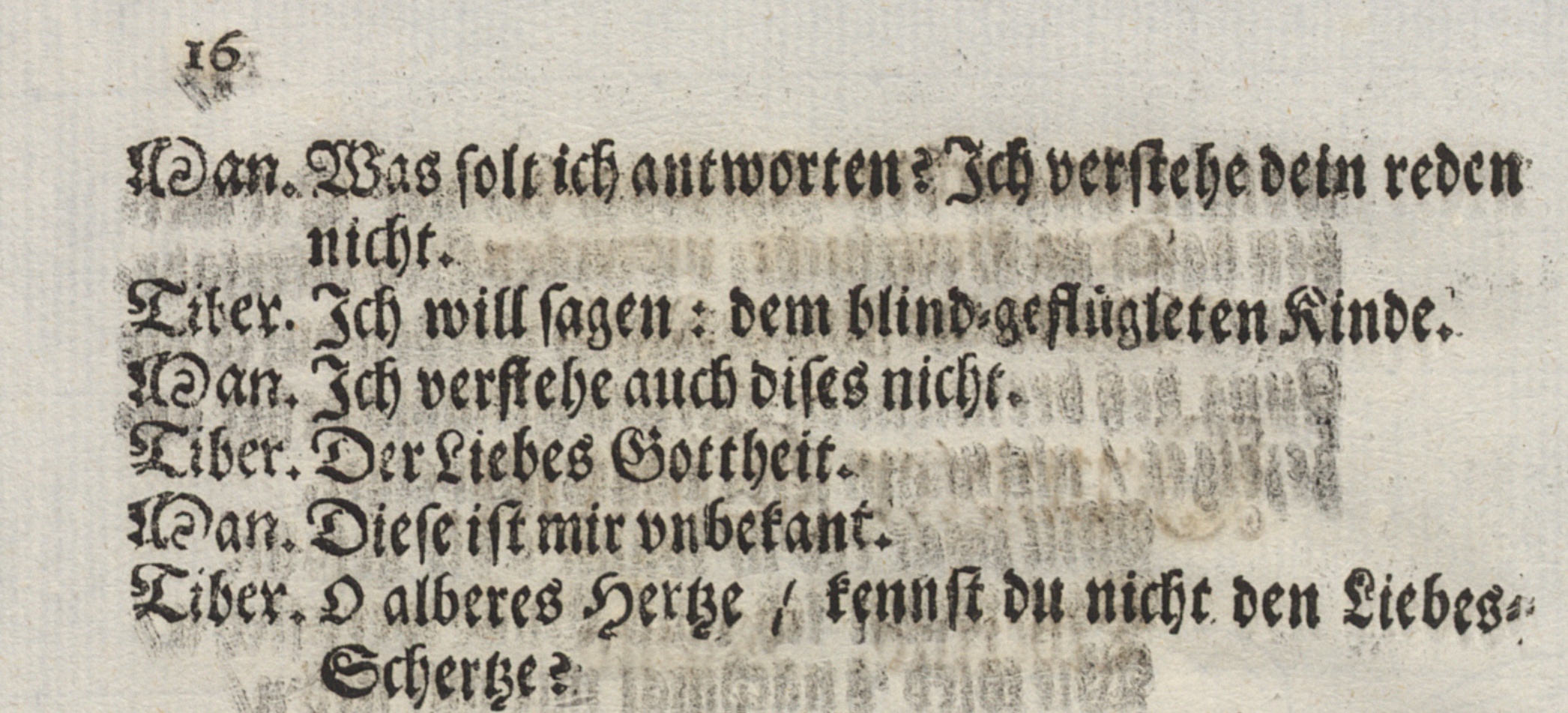}}&
\fbox{\includegraphics[width=.22\textwidth]{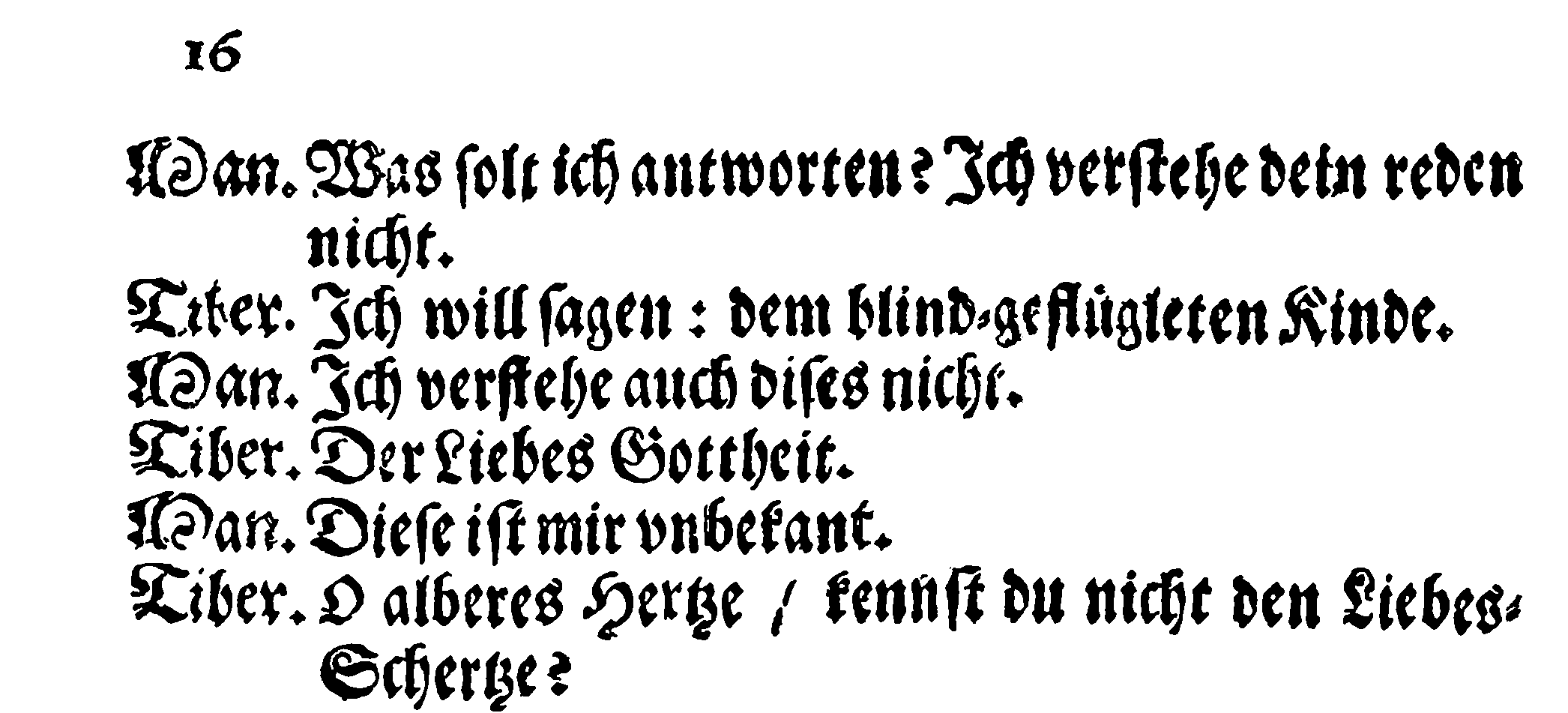}}&
\fbox{\includegraphics[width=.22\textwidth]{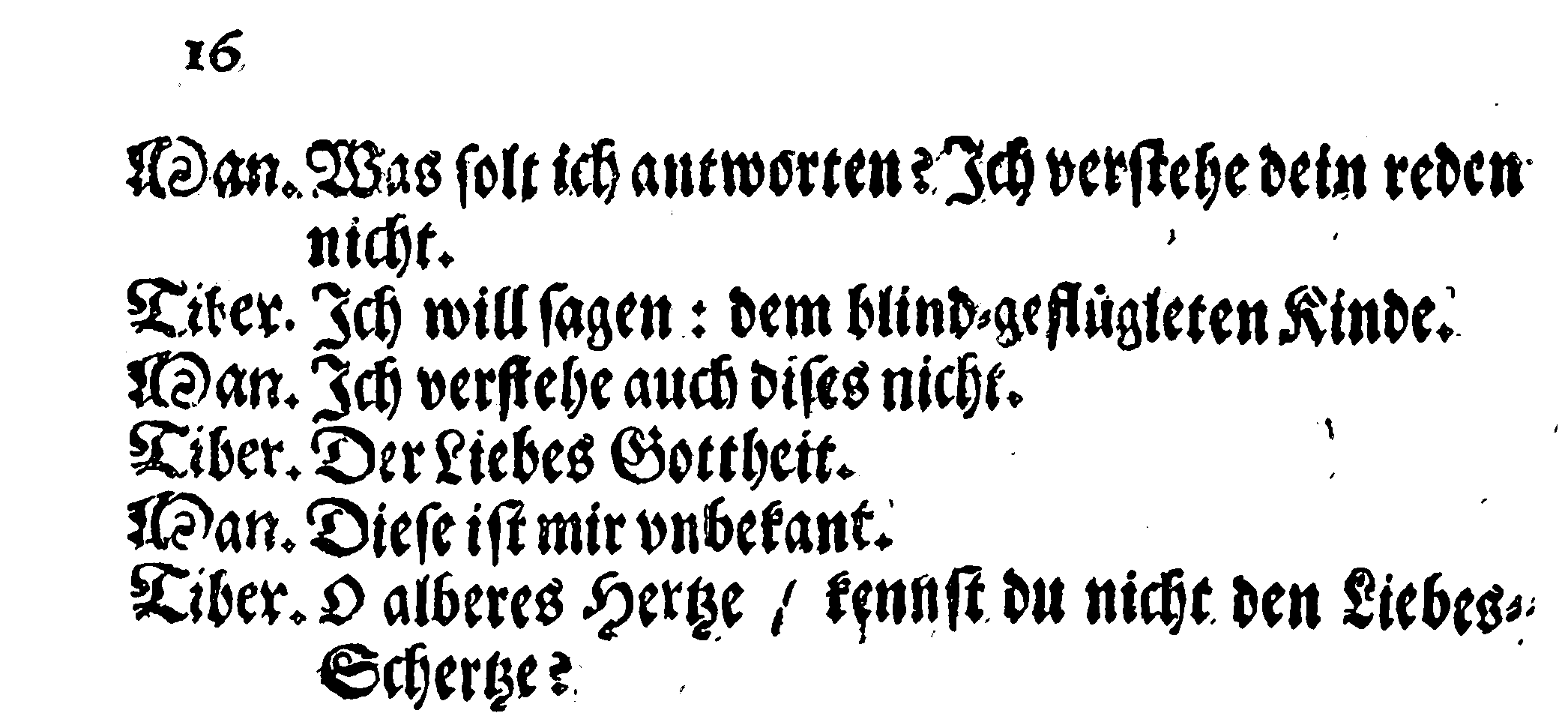}}&
\fbox{\includegraphics[width=.22\textwidth]{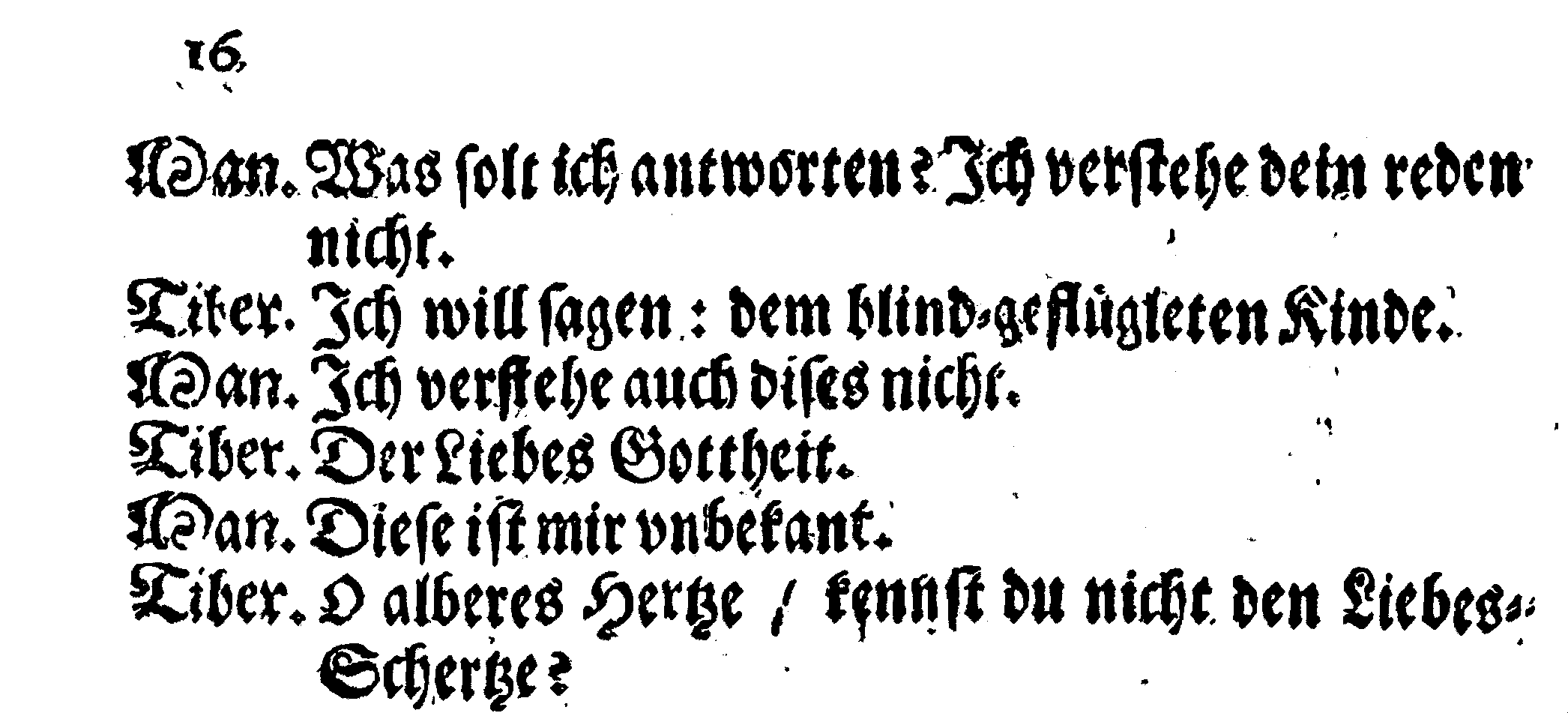}}\\
input & ground truth & SkipNetModel~\cite{Skip-Connected_ICPR18}  & M-32\\
\end{tabular}

\caption{Examples from DIBCO17~\cite{DIBCO17}. }
\label{fig:dibco17_example}
\end{figure}

We begin by discussing our results on document image binarization task. 
%During this experiment, we compared the proposed methods with state of the art techniques for document image binarization. 
For our experiment, we have considered the publicly available binarization dataset DIBCO13~\cite{DIBCO13} and DIBCO17~\cite{DIBCO17} as test sets. 
The proposed models and SkipNetModel are trained on the datasets~\cite{DIBCO09}, \cite{DIBCO10}, \cite{DIBCO11}, \cite{DIBCO12}, \cite{DIBCO14}, \cite{DIBCO16} and~\cite{DIBCO18}. 
While training the models for the test set DIBCO13~\cite{DIBCO13}, we also include the dataset DIBCO17~\cite{DIBCO17} into our training data. 
The models for this task are trained using the augmentation strategy described in Sec.~\ref{sec:result}. 
The same training strategy is also followed while training the models for the task DIBCO17~\cite{DIBCO17}. 
The models are compared using the DIBCO13~\cite{DIBCO13} evaluation criteria: F-measure, pseudo F-measure ($F_{ps}$), peak signal to noise ratio (PSNR), and distance reciprocal distortion (DRD). 
For the metrics F-measue, $F_{ps}$, and PSNR, higher values correspond to better performance whereas, in case of the metric DRD lower is better.
While evaluating our methods on DIBCO13 dataset, we have compared our methods with traditional binarization algorithms~\cite{otsu,Sauvola00adaptivedocument}, state of the art binarization techniques~\cite{Tensmeyer_Binary_ICDAR17,Vo_bin_PR18}, DE-GAN~\cite{DEGAN_TPAMI20} and SkipNetModel~\cite{Skip-Connected_ICPR18}. 
Overall performance of these methods are reported in Table~\ref{tab:dibco13}. 
In this table, performance of the methods~\cite{otsu,Sauvola00adaptivedocument,DEGAN_TPAMI20,Tensmeyer_Binary_ICDAR17,Vo_bin_PR18} are reported as they are reported in~\cite{DEGAN_TPAMI20}. 
From this table, it is evident that DE-GAN outperforms all other methods in terms of all metrics. 
However, the proposed method performs better than the traditional binarization algorithms~\cite{otsu,Sauvola00adaptivedocument} and they perform more or less similar to other state of the art techniques. 
Moreover, from Tables~\ref{tab:ops_params} and~\ref{tab:dibco13}, we can observe that though the proposed method can not outperform the state of the art techniques but they perform similar to most of the state of the art techniques with much lesser computational and memory cost. 
We have also reported the performance of the proposed methods with the top 5 methods of DIBCO17 competition~\cite{DIBCO17}, SkipNetModel and DE-GAN in Table~\ref{tab:dibco17}. 
A similar performance of the proposed methods is also observed from this table in comparison to the state of the art techniques. 
Typical examples from the datasets DIBCO13 and DIBCO17 are shown in Figs.~\ref{fig:dibco13_example} and~\ref{fig:dibco17_example}.

\begin{table}
\begin{center}
\setlength{\tabcolsep}{4pt}
\begin{tabular}{lcccccc}
\toprule
Model & F-measure & $F_{ps}$ & PSNR & DRD \\ %\hline
\midrule
10~\cite{DIBCO17} & $91.04$ & $92.86$ & $18.28$ & $3.40$ \\ %\hline
17a~\cite{DIBCO17} & $89.67$ & $91.03$ & $17.58$ & $4.35$ \\ %\hline
12~\cite{DIBCO17} & $89.42$ & $91.52$ & $17.61$ & $3.56$ \\ %\hline
1b~\cite{DIBCO17} & $86.05$ & $90.25$ & $17.53$ & $4.52$ \\ %\hline
1a~\cite{DIBCO17} & $83.76$ & $90.35$ & $17.07$ & $4.33$ \\ %\hline \hline
DE-GAN~\cite{DEGAN_TPAMI20} & $97.91$ & $98.23$ & $18.74$ & $3.01$ \\ %\hline
SkipNetModel~\cite{Skip-Connected_ICPR18} & $91.13$ & $92.91$ & $18.01$ & $3.22$ \\ %\hline
M-64 (proposed) & $90.80$ & $91.73$ & $17.84$ & $3.32$ \\ %\hline
M-32 (proposed) & $89.93$ & $90.61$ & $17.32$ & $3.74$ \\ %\hline
M-16 (proposed) & $87.81$ & $89.40$ & $16.91$ & $4.15$ \\ %\hline
\bottomrule
\end{tabular}
\end{center}
\caption{Results on DIBCO17~\cite{DIBCO17}. Here 10, 17a, 12, 1b, and 1a are the top 5 methods from DIBCO 2017 competition~\cite{DIBCO17}}
\label{tab:dibco17}
\end{table}

% \begin{table}
% \begin{center}
% \begin{tabular}{|c|c|c|c|c|c|c|}
% \hline
% Model & F-measure & $F_{ps}$ & PSNR & DRD \\ \hline
% 10~\cite{DIBCO17} & $91.04$ & $92.86$ & $18.28$ & $3.40$ \\ \hline
% 17a~\cite{DIBCO17} & $89.67$ & $91.03$ & $17.58$ & $4.35$ \\ \hline
% 12~\cite{DIBCO17} & $89.42$ & $91.52$ & $17.61$ & $3.56$ \\ \hline
% 1b~\cite{DIBCO17} & $86.05$ & $90.25$ & $17.53$ & $4.52$ \\ \hline
% 1a~\cite{DIBCO17} & $83.76$ & $90.35$ & $17.07$ & $4.33$ \\ \hline \hline
% DE-GAN~\cite{DEGAN_TPAMI20} & $97.91$ & $98.23$ & $18.74$ & $3.01$ \\ \hline
% SkipNetModel~\cite{Skip-Connected_ICPR18} & $91.13$ & $92.91$ & $18.01$ & $3.22$ \\ \hline
% M-64  & $90.80$ & $91.73$ & $17.84$ & $3.32$ \\ \hline
% M-32  & $89.93$ & $90.61$ & $17.32$ & $3.74$ \\ \hline
% M-16  & $87.81$ & $89.40$ & $16.91$ & $4.15$ \\ \hline
% \end{tabular}
% \end{center}
% \caption{Results on DIBCO17~\cite{DIBCO17}; Here 10, 17a, 12, 1b, 1a are the top 5 methods from DIBCO 2017 competition~\cite{DIBCO17}}
% \label{tab:dibco17}
% \end{table}

\begin{figure}[h!]\center

\begin{tabular}{@{}c@{\ }c@{\ }c@{\ }c@{}}
\fbox{\includegraphics[width=.22\textwidth]{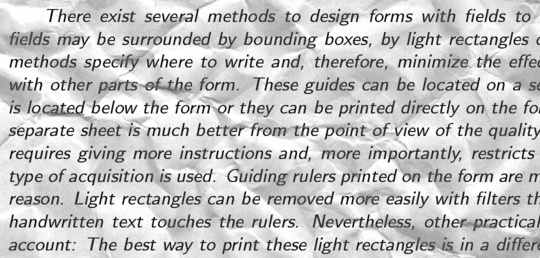}}&
\fbox{\includegraphics[width=.22\textwidth]{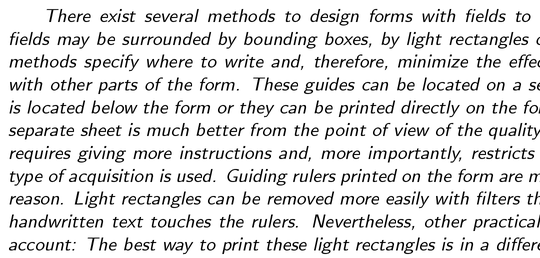}}&
\fbox{\includegraphics[width=.22\textwidth]{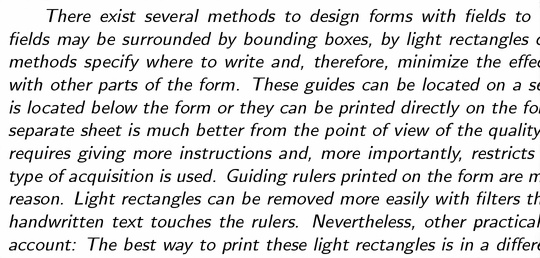}}&
\fbox{\includegraphics[width=.22\textwidth]{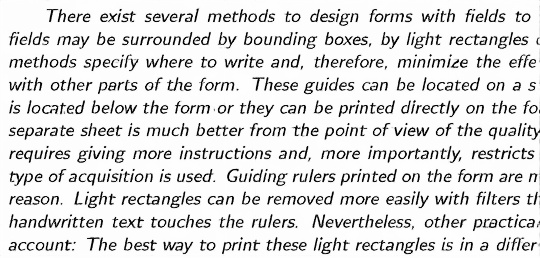}}\\
%\multicolumn{3}{c}{(a)} \\
%(a)&(b)&(c)
%\fbox{\includegraphics[width=.22\textwidth]{./images/noisyoffice/synthetic/image_179_in.jpeg}}&
%\fbox{\includegraphics[width=.22\textwidth]{./images/noisyoffice/synthetic/clean_image_179.png}}&
%\fbox{\includegraphics[width=.22\textwidth]{./images/noisyoffice/synthetic/image_179_skip.jpeg}}&
%\fbox{\includegraphics[width=.22\textwidth]{./images/noisyoffice/synthetic/image_179_dnn_DIBCO18Model.jpeg}}\\
\fbox{\includegraphics[width=.22\textwidth]{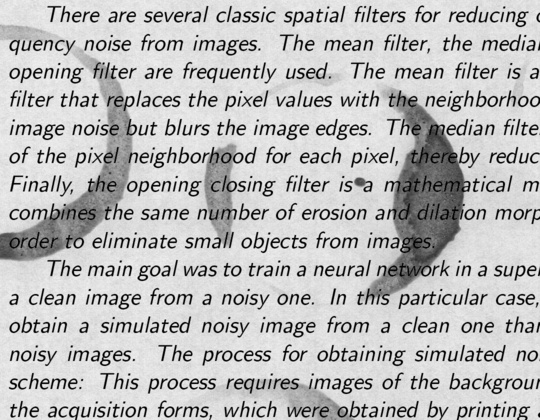}}&
\fbox{\includegraphics[width=.22\textwidth]{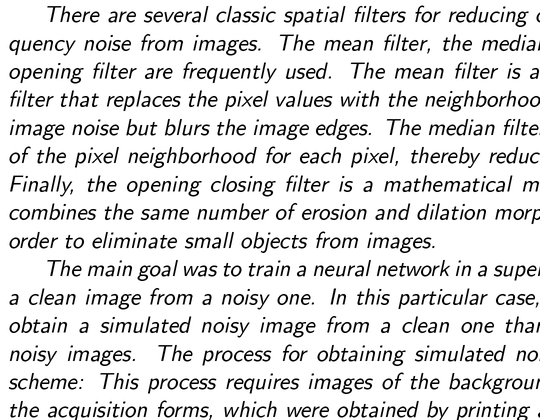}}&
\fbox{\includegraphics[width=.22\textwidth]{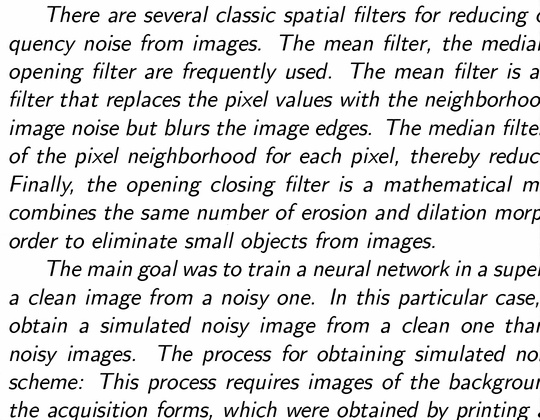}}&
\fbox{\includegraphics[width=.22\textwidth]{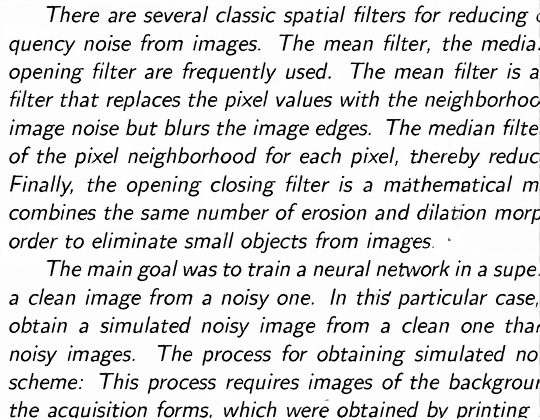}}\\
input & ground truth & SkipNetModel~\cite{Skip-Connected_ICPR18}  & M-32\\
\end{tabular}
\caption{Typical examples of noisy images from our test set of synthetic data from  \cite{NoisyOffice}. }
\label{fig:noisyoffice_example}
\end{figure}

\begin{figure}[h!]\center

\begin{tabular}{@{}c@{\ }c@{\ }c@{}}
\fbox{\includegraphics[width=.3\textwidth]{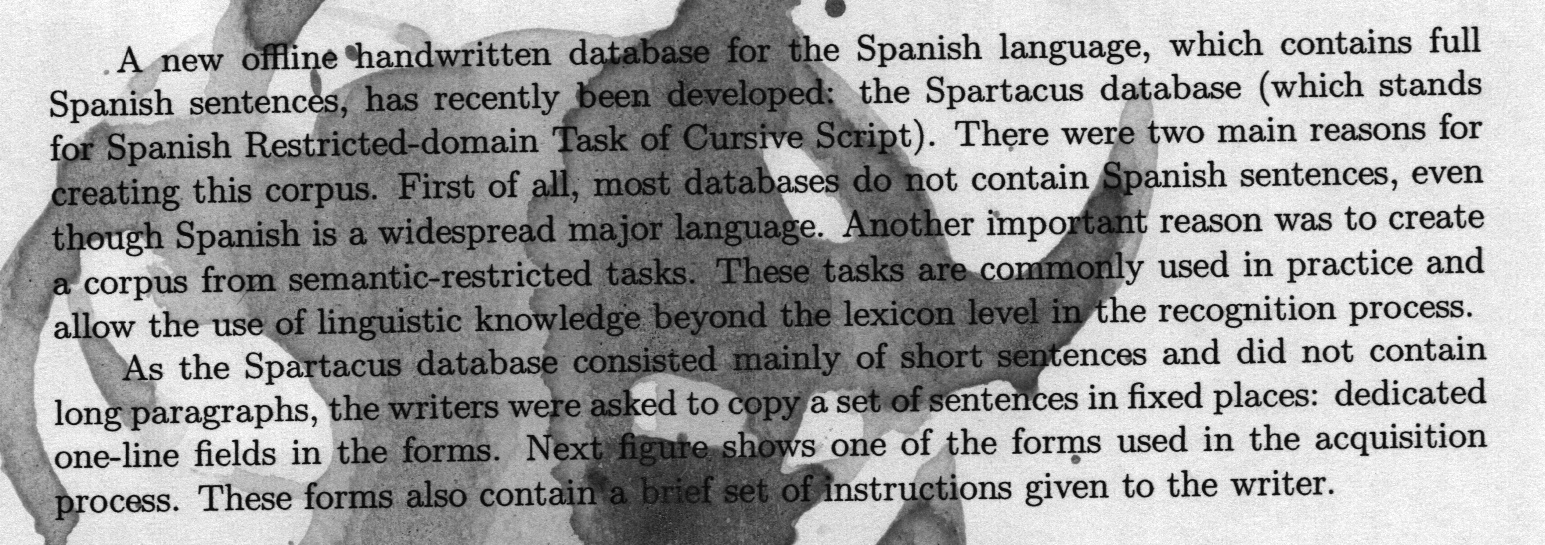}}&
\fbox{\includegraphics[width=.3\textwidth]{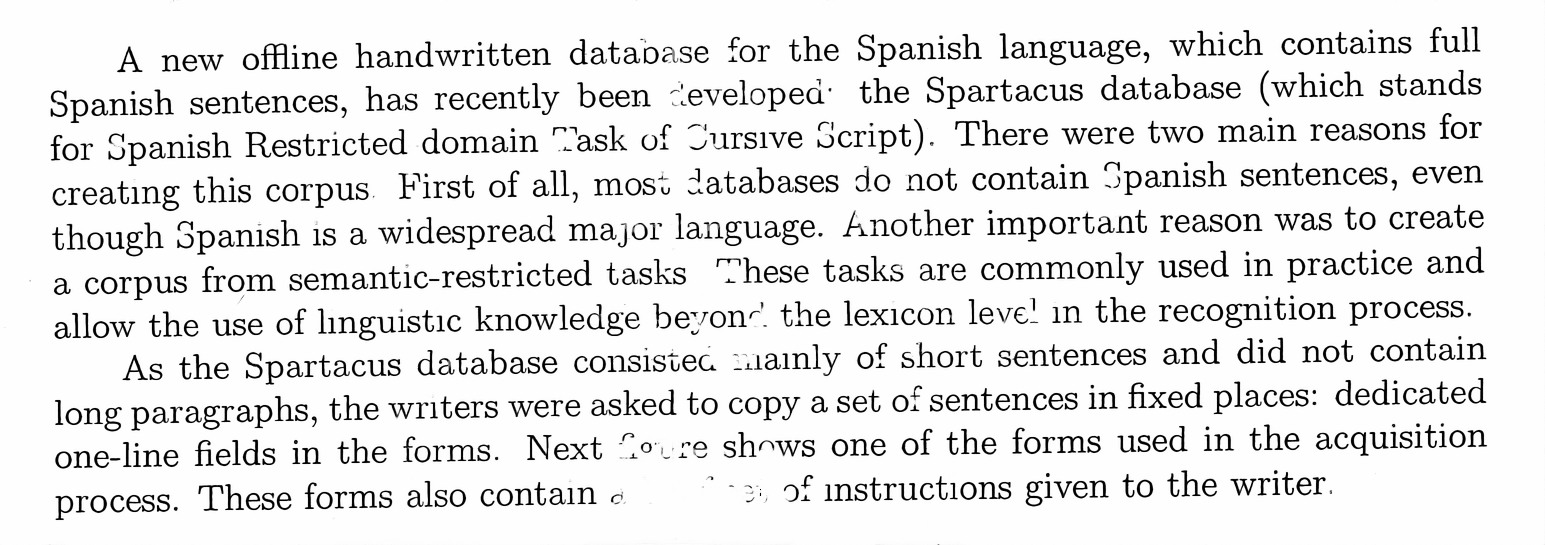}}&
\fbox{\includegraphics[width=.3\textwidth]{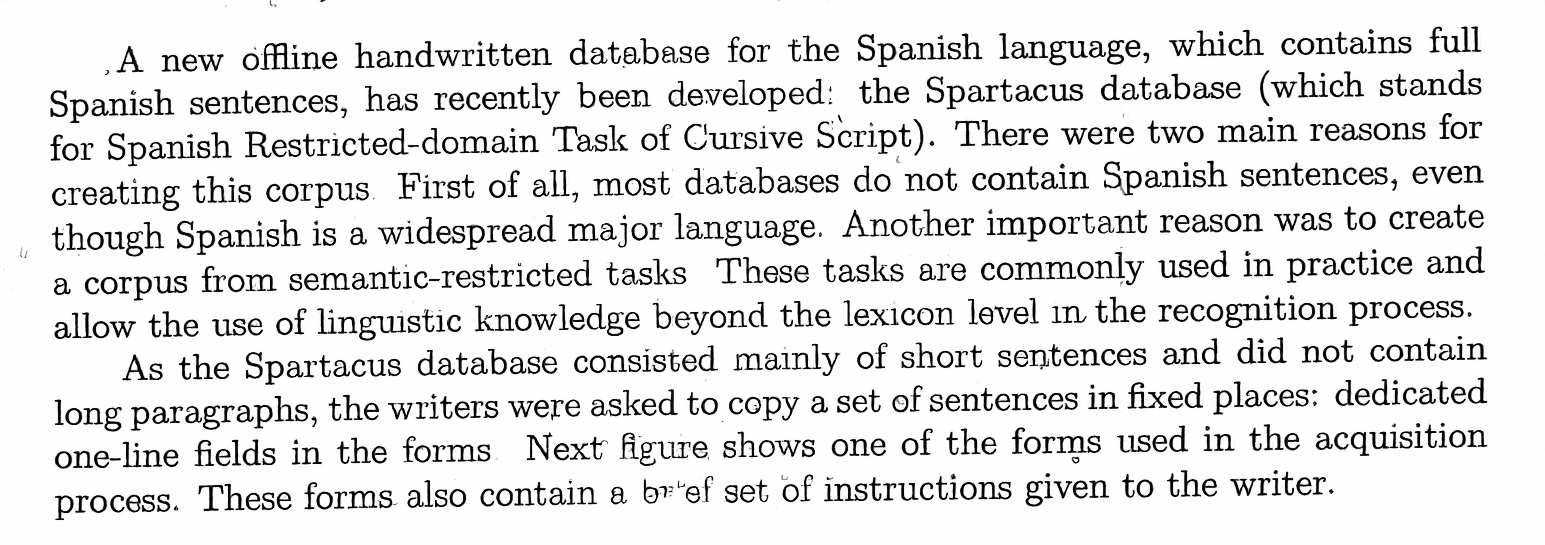}}\\

%\fbox{\includegraphics[width=.3\textwidth]{./images/noisyoffice/real/Fontnre_Noisec_RE_in.jpeg}}&
%\fbox{\includegraphics[width=.3\textwidth]{./images/noisyoffice/real/Fontnre_Noisec_RE_skip.png}}&
%\fbox{\includegraphics[width=.3\textwidth]{./images/noisyoffice/real/Fontnre_Noisec_RE_illu.png}}\\

\fbox{\includegraphics[width=.3\textwidth]{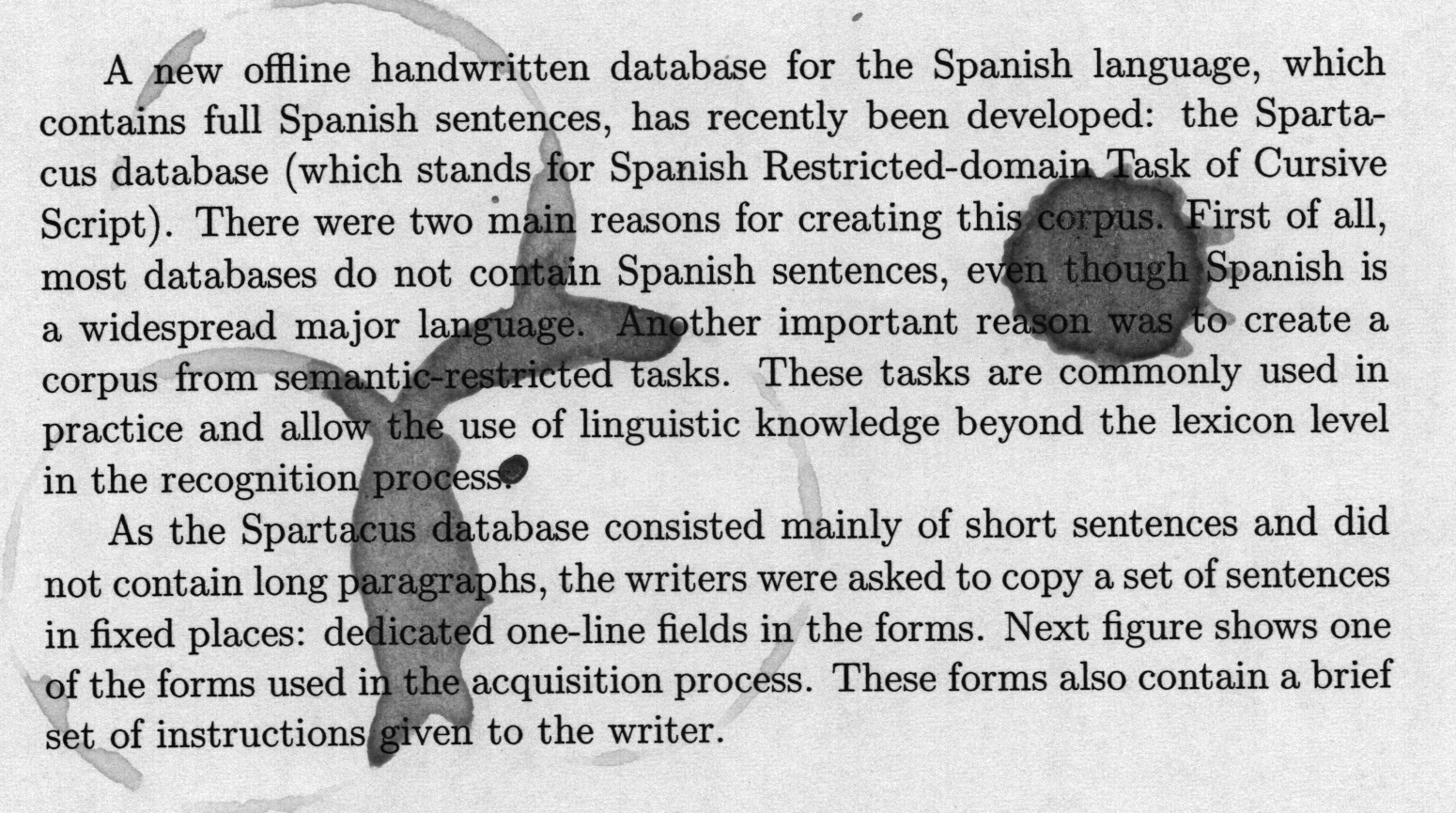}}&
\fbox{\includegraphics[width=.3\textwidth]{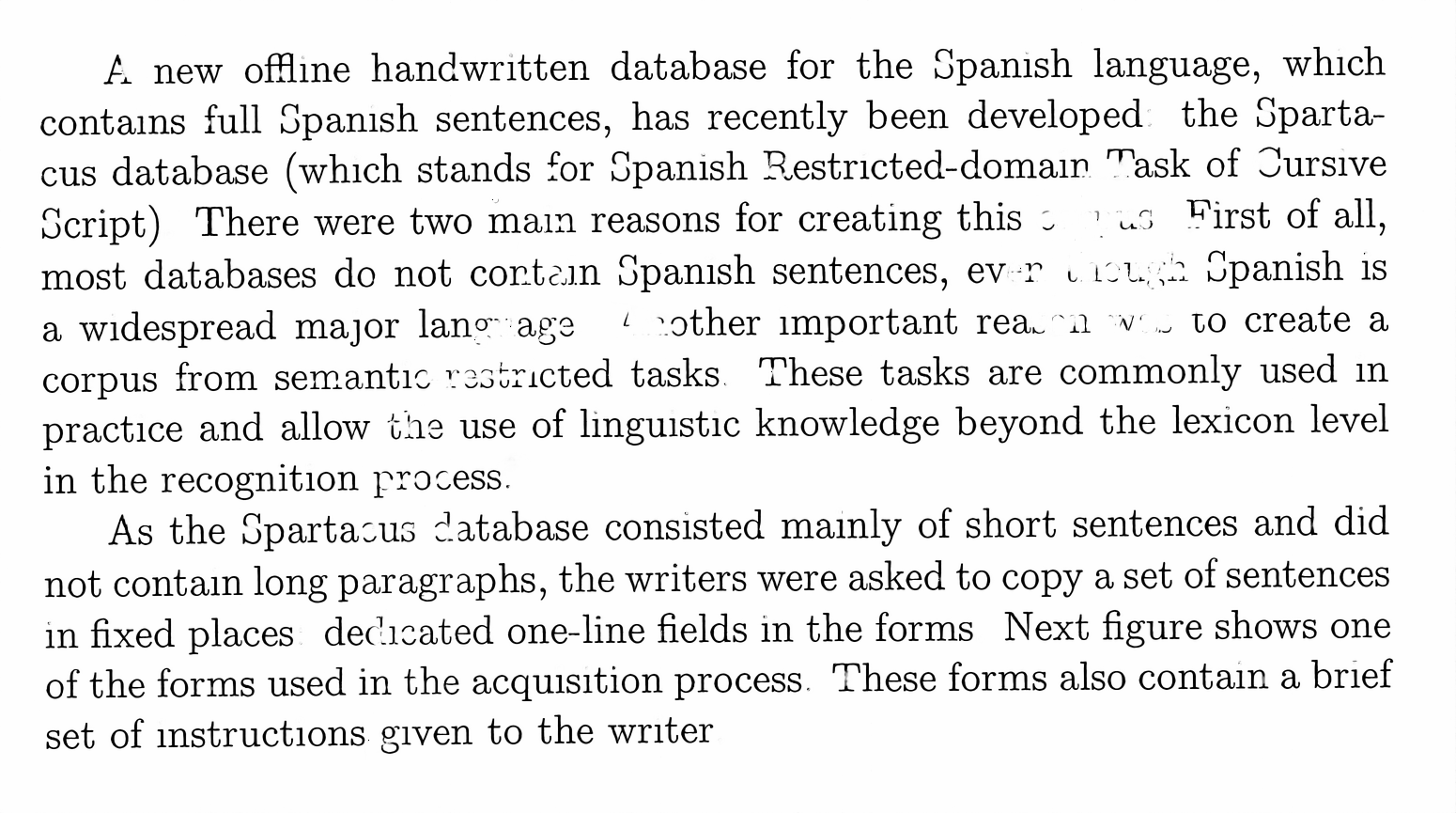}}&
\fbox{\includegraphics[width=.3\textwidth]{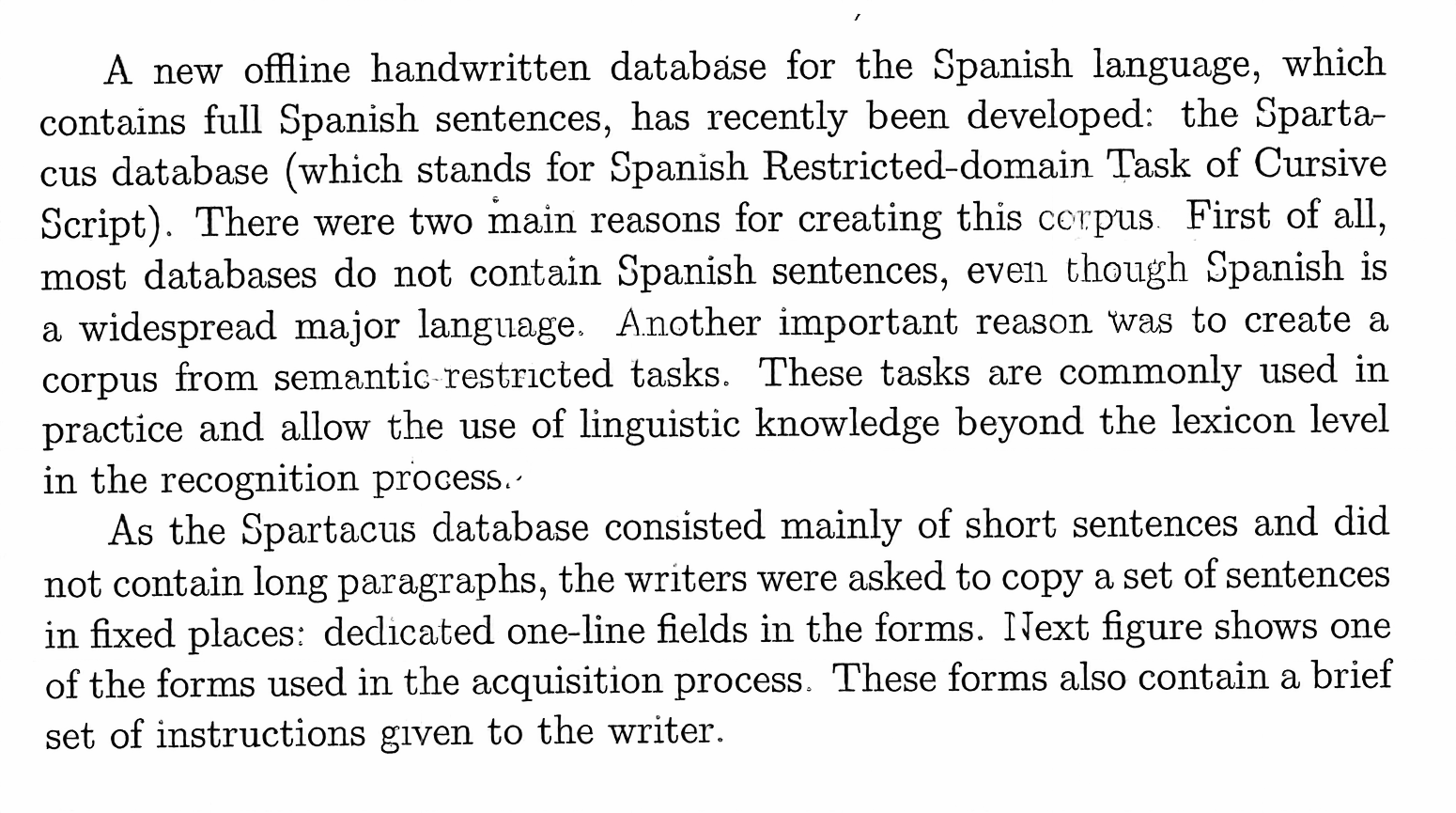}}\\

input & SkipNetModel~\cite{Skip-Connected_ICPR18}  & M-32\\

\end{tabular}
\caption{Typical examples of noisy images of real data from~\cite{NoisyOffice}. }
\label{fig:noisyoffice_real_example}
\end{figure}

\subsection{Gray scale and color cleanup}

%\begin{figure}[h!]
% \centering
% \begin{tabular}{@{}c@{\ }c@{}}
%\fbox{\includegraphics[width=.4\textwidth]{./images/PSNR_varydata1.png}}&
%\fbox{\includegraphics[width=.4\textwidth]{./images/SSIM_varydata1.png}}\\

%PSNR & SSIM \\

%\end{tabular}
 % LayoutTypes.pdf: 595x842 pixel, 72dpi, 20.99x29.70 cm, bb=0 0 595 842
% \caption{PSNR and SSIM scores of the models on NoisyOffice dataset~\cite{NoisyOffice} with varying the training data from $25 \%$ to $100 \%$}
% \label{fig:PSNR_perf}
%\end{figure}

\begin{figure}[t]
 \centering

\fbox{\includegraphics[width=.75\textwidth]{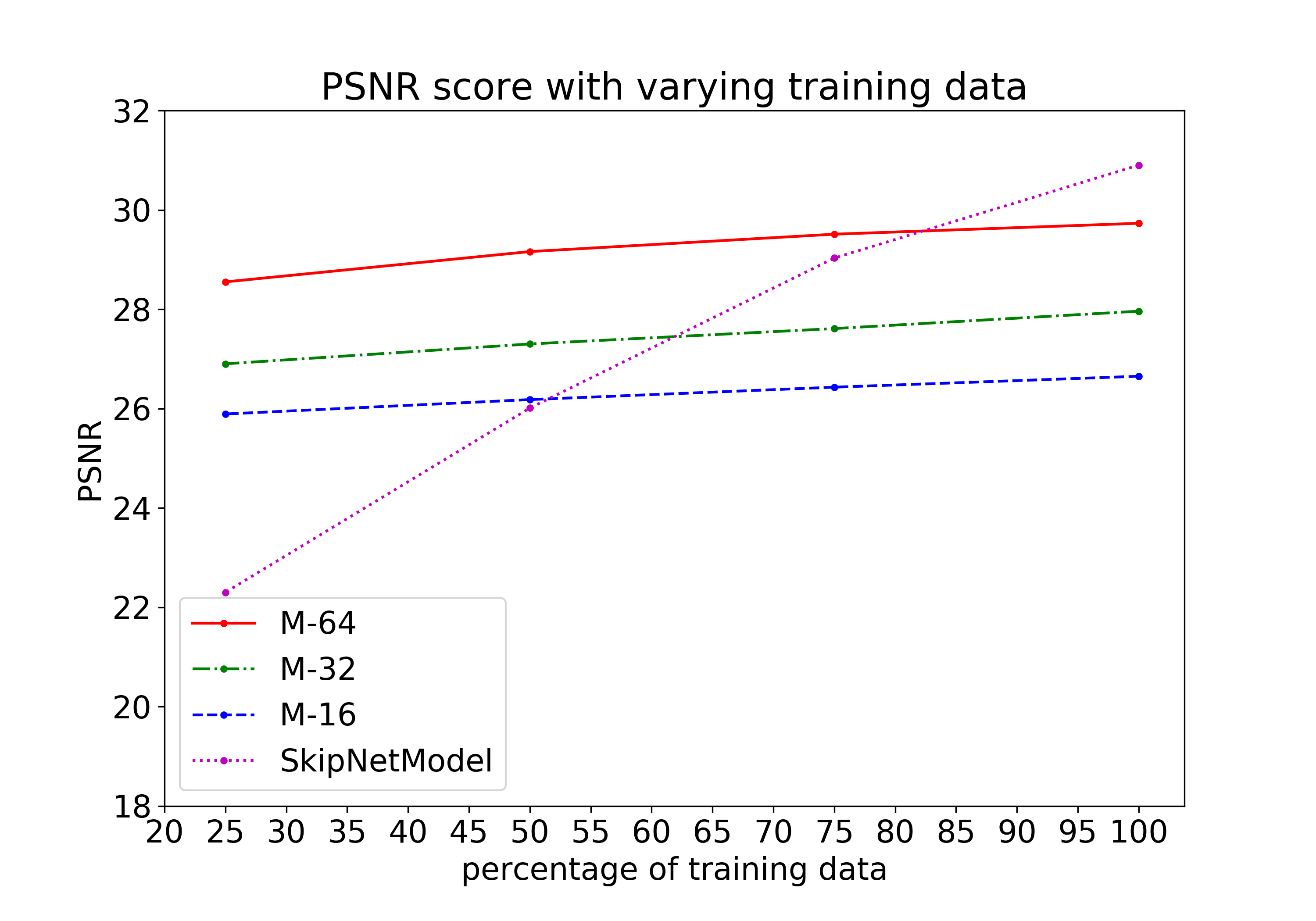}}

 % LayoutTypes.pdf: 595x842 pixel, 72dpi, 20.99x29.70 cm, bb=0 0 595 842
 \caption{PSNR scores of the models on NoisyOffice dataset~\cite{NoisyOffice} with varying the training data from $25 \%$ to $100 \%$}
 \label{fig:PSNR_perf}
\end{figure}

For the purpose of document cleanup, we first show the effectiveness of the proposed methods in gray scale. 
\textbf{The gray scale cleanup} part of our experiment is conducted on the publicly available dataset {\textit NoisyOffice}~\cite{Dua:2019,NoisyOffice} . 
This dataset consists of two parts, first one is real noisy images consisting of $72$ files, and a synthetic dataset consisting of $216$ files. 
There were no groundtruth images available for the real data, therefore, we are not able to include the real dataset for quantitative analysis of our experiment. 
The model is trained and evaluated on the synthetic data. 
We divide the synthetic data into two parts - $172$ images for training and $44$ images for testing.
The authors of~\cite{DEGAN_TPAMI20} did not share the saved model for gray scale cleanup. 
Therefore, for this experiment, only the method proposed in ~\cite{Skip-Connected_ICPR18} is used for comparison. 
To measure the capability of the proposed models with respect to removing noise, we adopt peak signal to noise ratio (PSNR) as the quality metric. 
%and structural similarity index (SSIM) 
In order to determine the dependence of the model performance on the amount of available training data,  
we have trained the SkipNetModel and the proposed models M-64, M-32, and M-16 by varying the amount of training data from $25 \%$ to $100 \%$. 
The performance of the models with respect to PSNR score is shown in Fig.~\ref{fig:PSNR_perf}. 
It can be observed from this figure that in presence of $100 \%$ training data, SkipNetModel~\cite{Skip-Connected_ICPR18} performs better than the proposed models. 
However, it can also be observed from this figure that the performance of the proposed models is more or less remains the same. 
It can also be observed from this figure that the performance of SkipNetModel varies a lot with the variation in the amount of training data. 
Typical examples of inputs, groundtruths from the test set along with the outputs of the models trained on $100 \%$ training data are shown in Fig.~\ref{fig:noisyoffice_example}. 
We have also shown a few examples of inputs and  outputs of the trained models on real data in Fig.~\ref{fig:noisyoffice_real_example}. 
It can be seen from this figure that the model M-32 performs better than SkipNetModel in few of the examples. 
From Figs.~\ref{fig:noisyoffice_real_example} and ~\ref{fig:PSNR_perf}, we can conclude that the proposed model is more generalized and performs more robustly with respect to SkipNetModel.

%\begin{comment}
%\begin{table}
%\label{tab:pr_trnd}
%\begin{center}
%\begin{tabular}{|c|c|c|c|c|}
%\hline
%Model &   $100 \%$ & $75 \%$ & $50 \%$ & $25 \%$\\ \hline
%SkipNetModel~\cite{Skip-Connected_ICPR18} & $0.995, 30.9$ & $0.993, 29.03$ & $0.99, 26.01$ &  $0.982, 22.3$ \\ \hline
%M-64  & $0.99, 29.73$ & $0.99, 29.51$  & $0.99, 29.16$  & $0.988, 28.55$ \\ \hline
%M-32  & $0.985, 27.96$ & $0.985, 27.61$  & $0.983, 27.3$  & $0.982, 26.9$\\ \hline
%M-16  & $0.98, 26.65$ & $0.98, 26.93$ & $ 0.978, 26.18$ & $ 0.974, 25.89$ \\ \hline
%\end{tabular}
%\end{center}

%\caption{PERF VARYING DATA}
%\end{table}

%\begin{table}
%\label{tab:noisy_gray}
%\begin{center}
%\begin{tabular}{|c|c|c|c|}
%\hline
%Model &   parameters & SSIM & PSNR \\ \hline
%SkipNetModel~\cite{Skip-Connected_ICPR18} & $\sim 2.1 M$ & $0.995$ & $30.9$ \\ \hline
%M-64  & $\sim 0.456 M$  & $0.99$ & $29.73$ \\ \hline
%M2  & $\sim 0.263 M$  & $0.987$ & $28.34$ \\ \hline
%M-32  & $\sim 0.113 M$  & $0.985$ & $27.96$ \\ \hline
%M4  & $\sim 0.077 M$  & $0.984$ & $27.31$ \\ \hline
%M-16  & $\sim 0.027 M$  & $0.983$ & $27.13$ \\ \hline

%\end{tabular}
%\end{center}

%\caption{Results on our test set of the synthetic data from  NoisyOffice dataset~\cite{NoisyOffice}}
%\end{table}
%\end{comment}

\begin{table}[b]
\begin{center}
\setlength{\tabcolsep}{4pt}
\begin{tabular}{lcc}
\toprule
Model  & SSIM & PSNR \\ %\hline
\midrule
%- & noisy input & $0.87$ & $16.25$ \\ \hline
M-64 (proposed) & $0.967$ & $22.8$ \\ %\hline
M-32 (proposed) & $0.950$ & $21.4$ \\ %\hline
M-16 (proposed) & $0.923$ & $19.8$ \\ %\hline
%SkipNetModel~\cite{Skip-Connected_ICPR18} & $859.26$ & $0.93$ & $20.9$ \\ \hline
%M-64  & $859.26$ & $0.93$ & $20.9$ \\ \hline
%M-16  & $859.26$ & $0.93$ & $20.9$ \\ \hline
\bottomrule
\end{tabular}
\end{center}
\caption{Color cleanup performance of the proposed model: SSIM and PSNR score of the noisy input images with respect to the groundtruth are $0.87$ and $16.3$ respectively.}
\label{tab:clr_cleanup}
\end{table}

%\subsection{Color Clean up} 

\begin{figure}[t]\center
\begin{tabular}{@{}c@{\ }c@{\ }c@{\ }c@{}}
\fbox{\includegraphics[width=.22\textwidth]{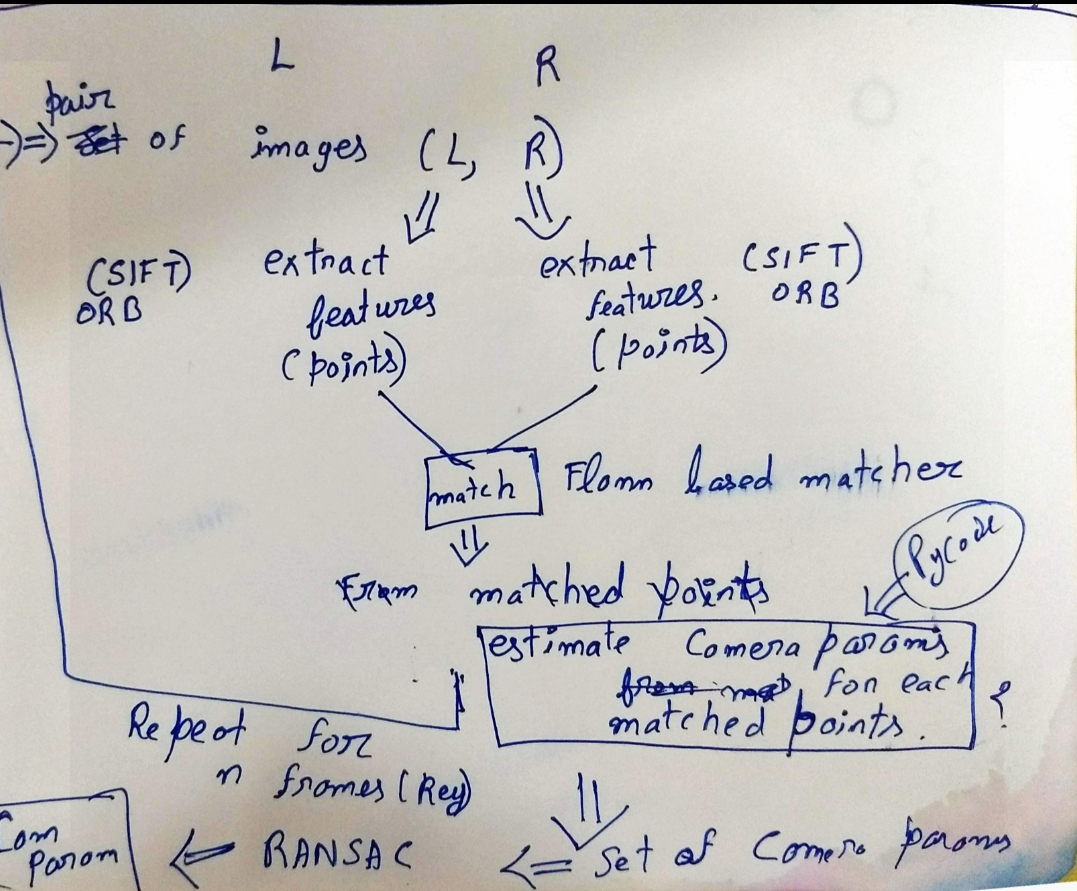}}&
\fbox{\includegraphics[width=.22\textwidth]{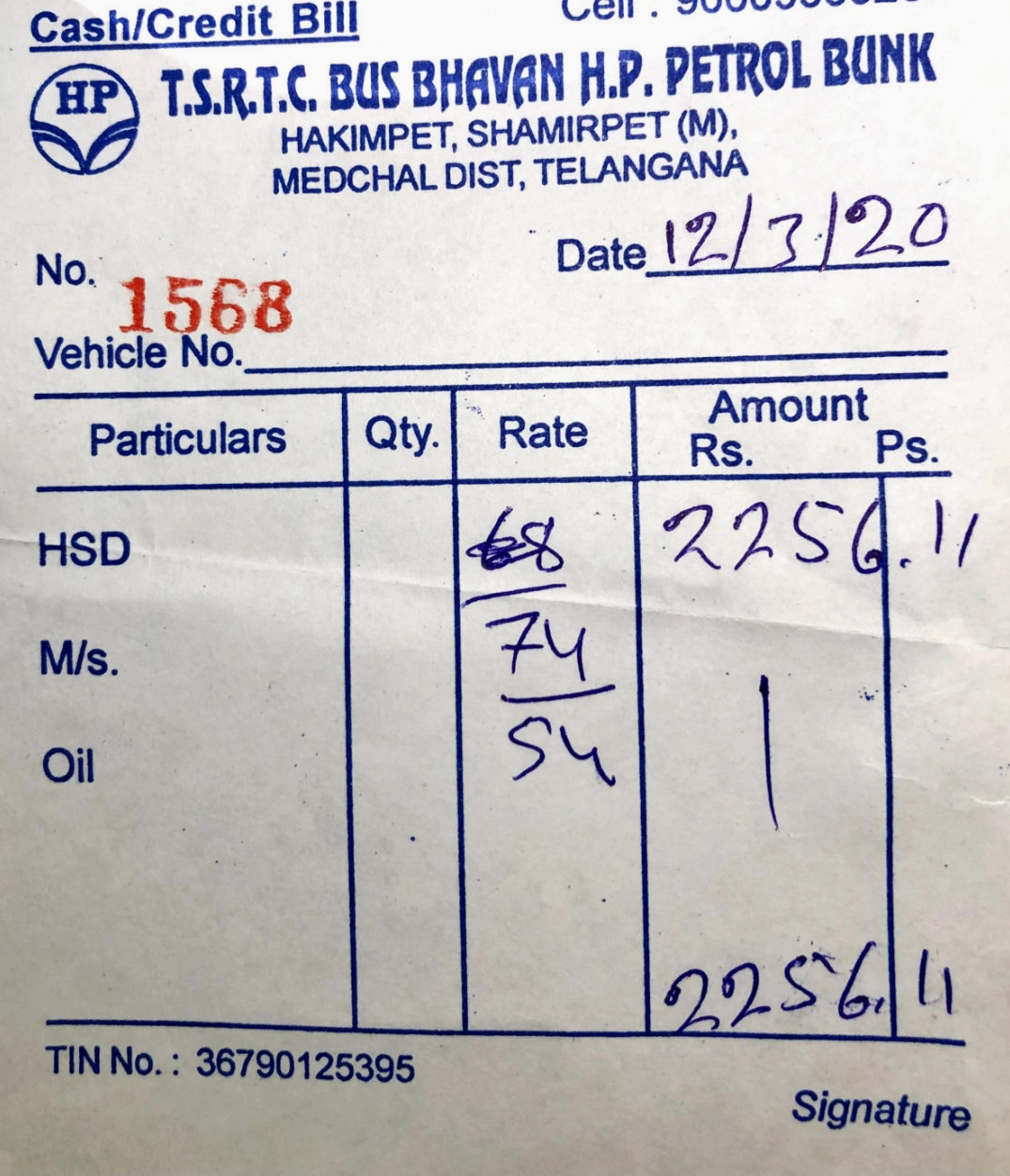}}&
\fbox{\includegraphics[width=.22\textwidth]{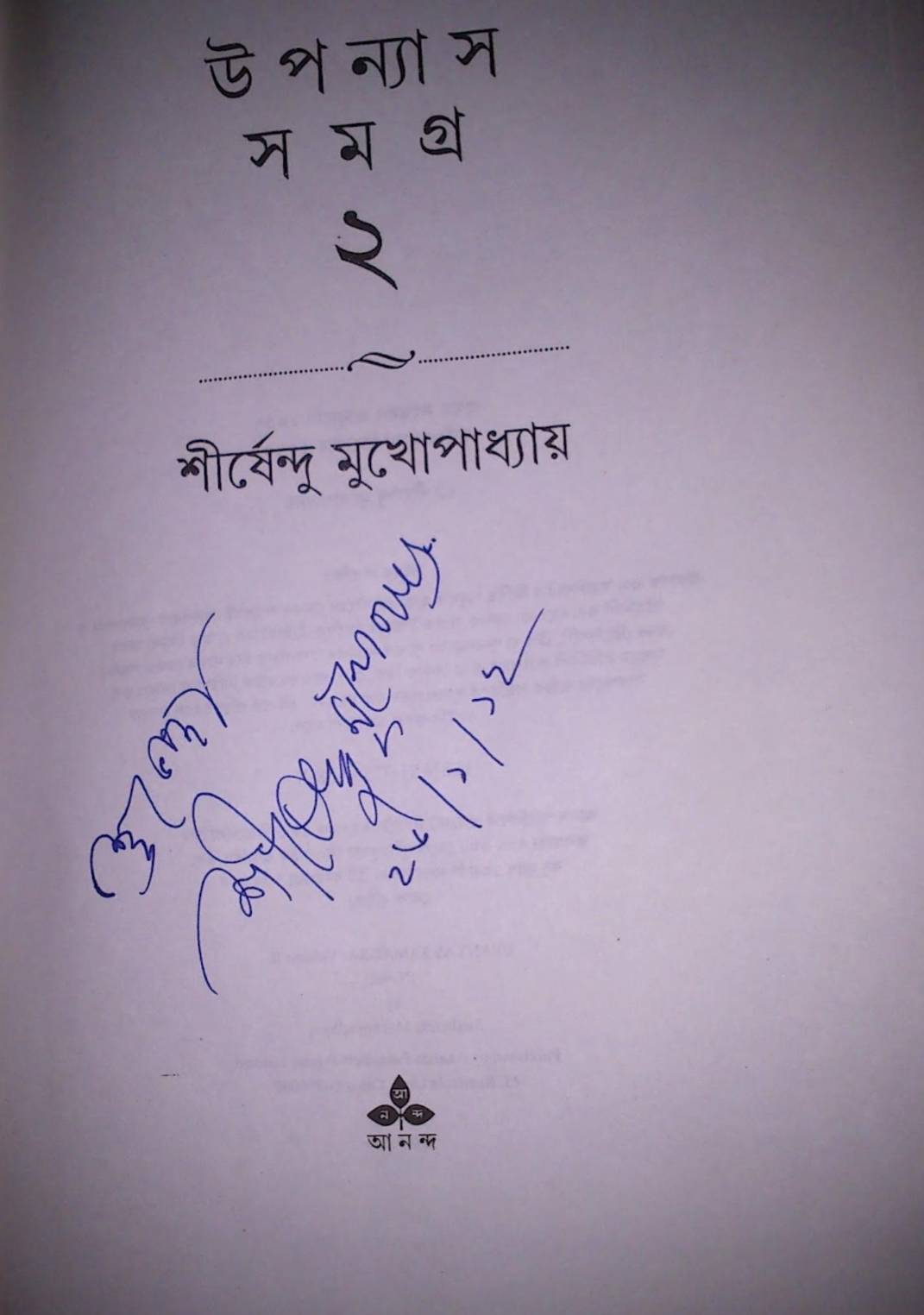}}&
\fbox{\includegraphics[width=.22\textwidth]{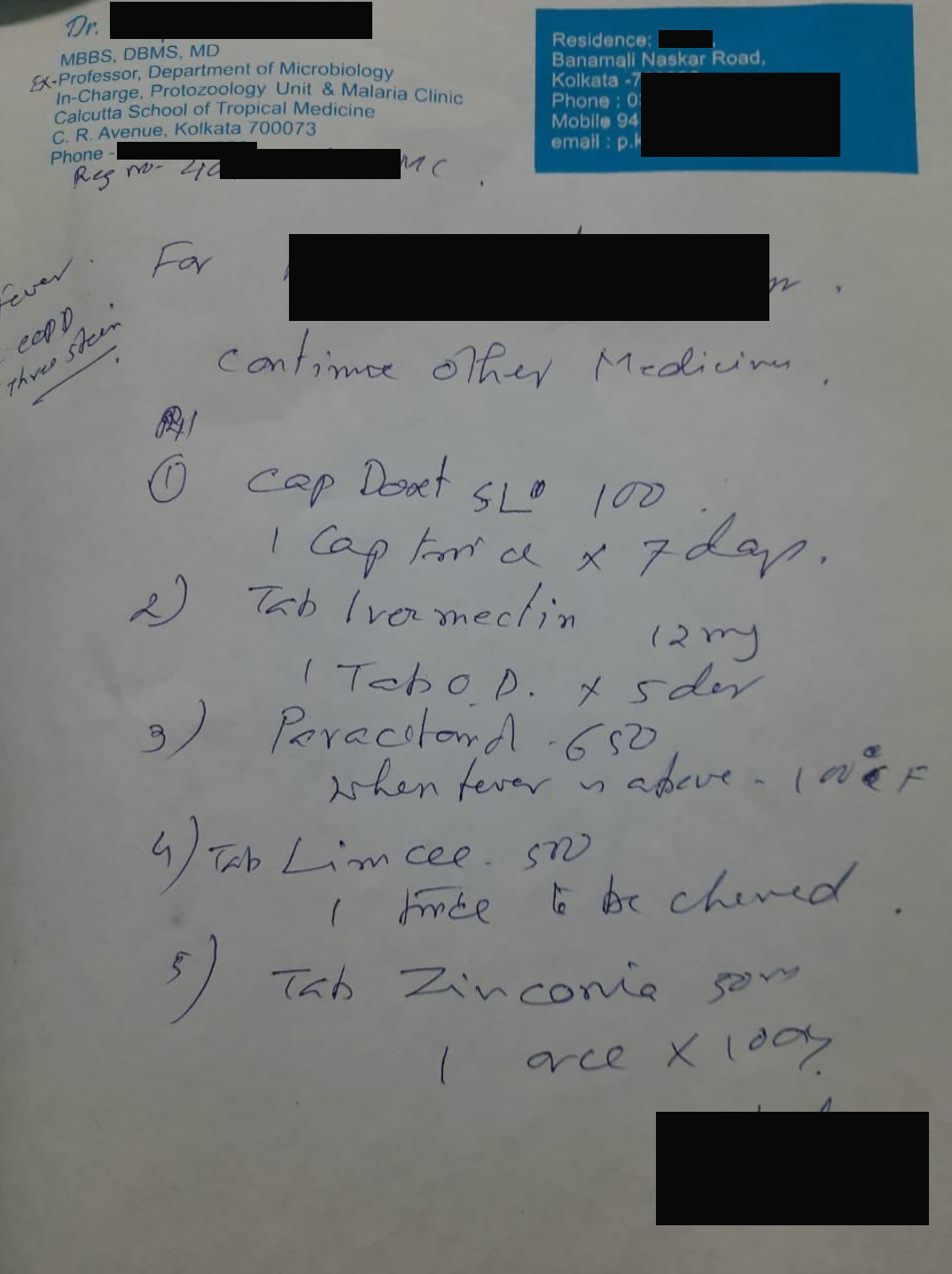}}\\
\multicolumn{4}{c}{(a)} \\
%(a)&(b)&(c)
\fbox{\includegraphics[width=.22\textwidth]{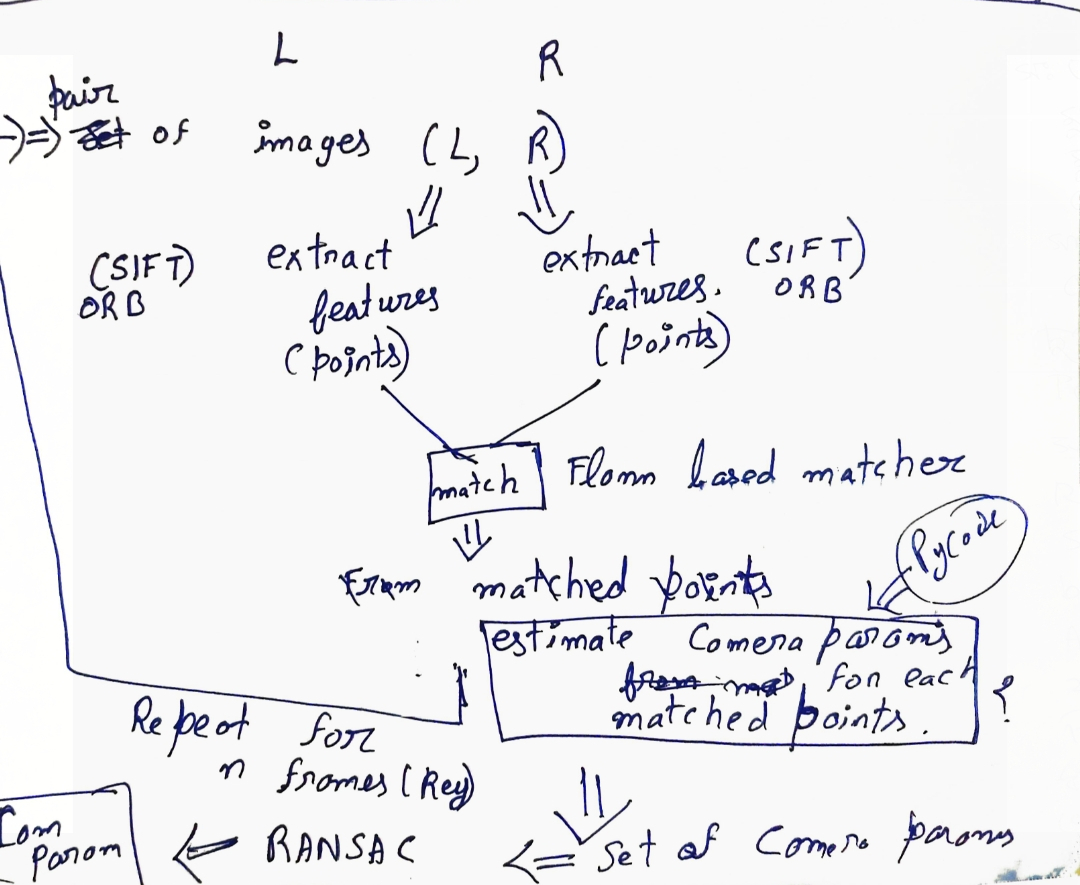}}&
\fbox{\includegraphics[width=.22\textwidth]{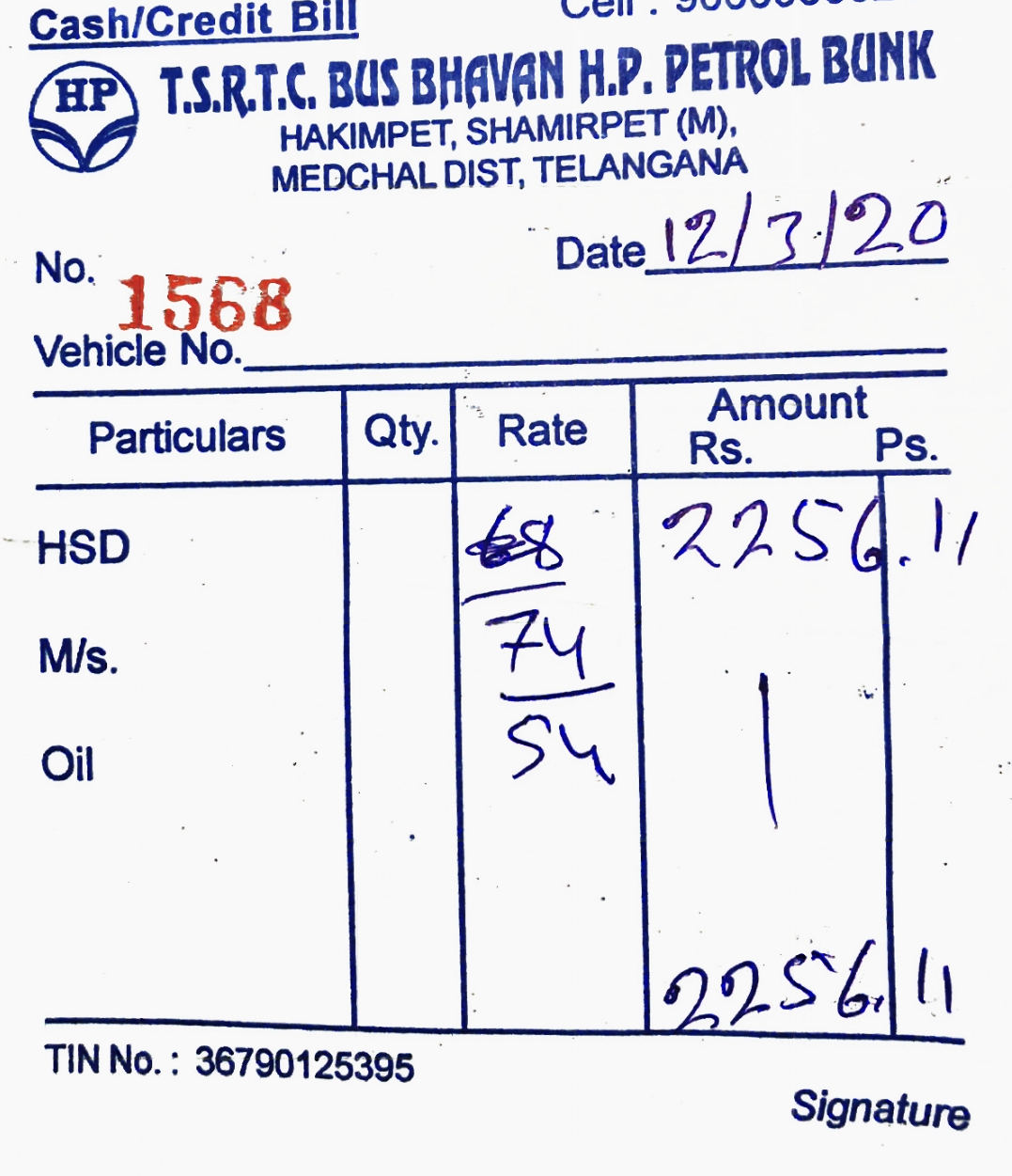}}&
\fbox{\includegraphics[width=.22\textwidth]{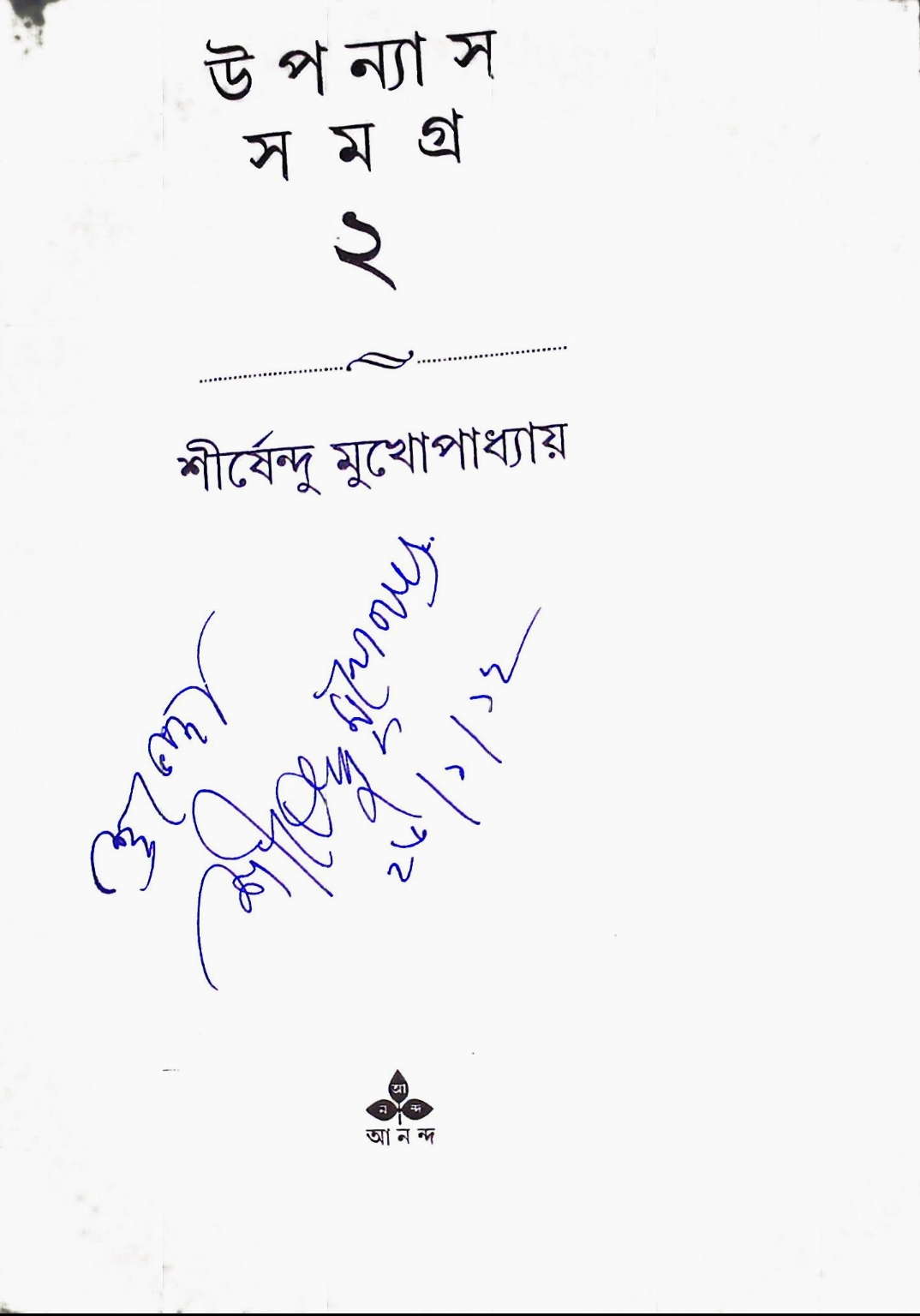}}&
\fbox{\includegraphics[width=.22\textwidth]{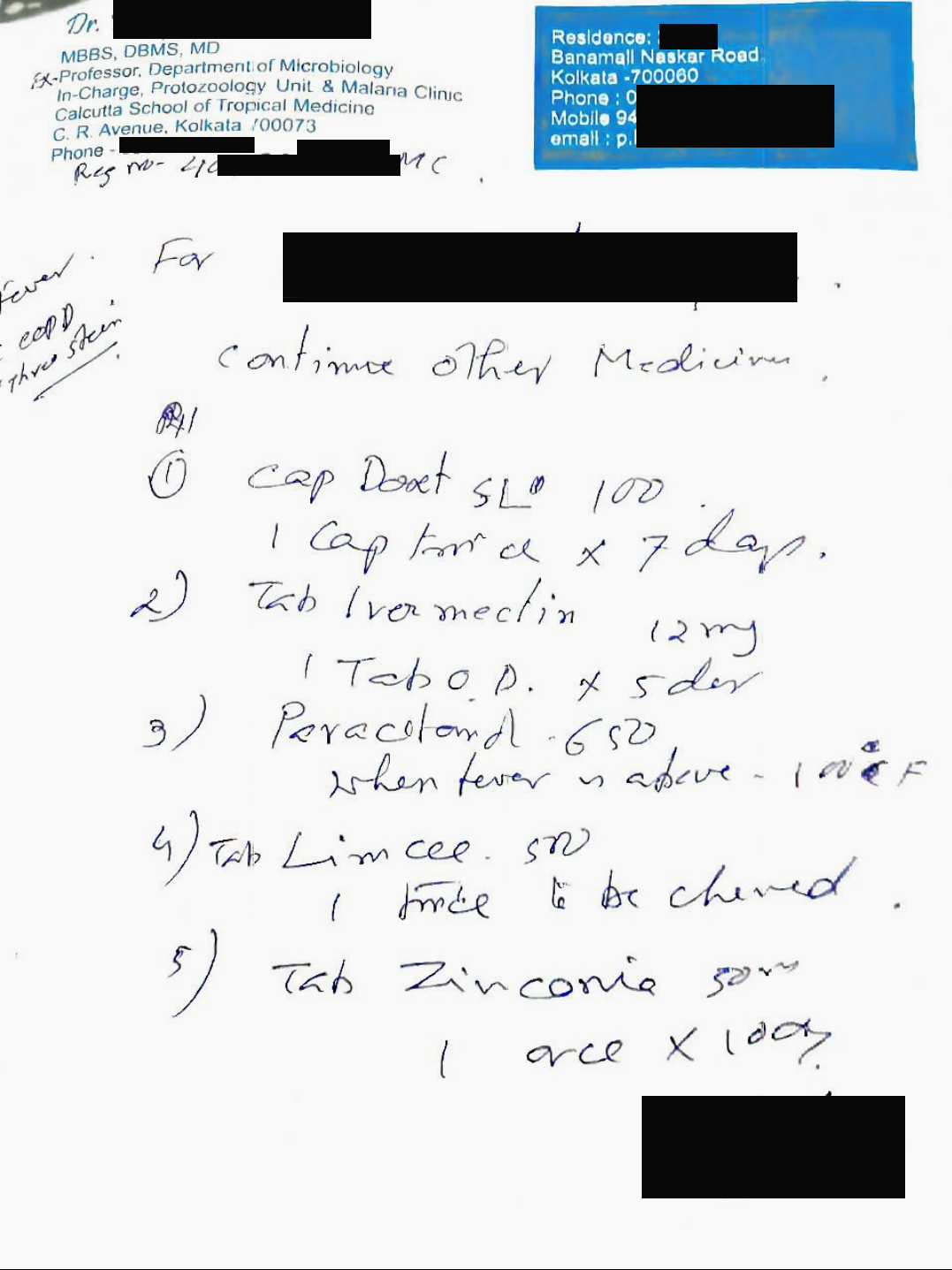}}\\
 \multicolumn{4}{c}{(b)}\\
\end{tabular}

\caption{Typical examples of color clean up. (a)~random inputs images; (b)~cleaned outputs using M-32 based model on a mobile device}
\label{fig:colorclean_example}
\end{figure}

Finally, we present our experimental results with respect to document \textbf{color cleanup} task. 
One challenging aspect of color cleanup is the preservation of color of the foreground pixels. 
For this experiment, we used a color dataset consisting of 250 mobile captured images. 
Each of the images are manually cleaned. 
We followed the same training strategy for training our model as described in Sec.~\ref{sec:result}. 
Random real life images (not belonging to the train/test set) and their corresponding outputs are shown in Fig.~\ref{fig:colorclean_example}. 
From this figure, we can observe a decent performance of the proposed model in performing color cleanup of document images. 
However, to provide a quantitative measure of our method, we compute PSNR, and structural similarity index (SSIM) score on the test set of the data in Table~\ref{tab:clr_cleanup}.

\section{Conclusion}
\label{sec:conclusion}

We have proposed an encoder-decoder based document cleanup model for resource constrained environments. To this end, we design a light-weight deep network with only a few residual blocks and skip connections. Our loss function incorporates the perceptual loss, which enables transfer learning from pre-trained deep CNN networks. 

We develop three models based on our network design, with varying network width. In terms of the number of parameters and product-sum operations, our models are 65-1030 and 3-27 times, respectively, smaller than a recently proposed GAN based document enhancement model~\cite{DEGAN_TPAMI20}. In spite of our relatively low network capacity, the generalization performance of our models on various benchmarks are encouraging and comparable with several document image cleanup techniques with deep architectures such as \cite{Tensmeyer_Binary_ICDAR17,Skip-Connected_ICPR18}. In addition, our models are more robust to low training data regime than \cite{Skip-Connected_ICPR18}. Hence, the proposed models offer a favorable trade-off between memory/latency and accuracy, making them suitable for mobile document image cleanup applications.

% Empirically, we observe that the perceptual loss allows the proposed network achieve a good generalization performance even with small training dataset. 
% In contrast existing deep architectures for document cleanup exhibit severe 
% To this end, we have designed a shallow network with less computational overhead to make the model suitable for memory and energy constrained devices. To achieve better learning capability in a shallow network, we have incorporated perceptual loss based transfer learning. 
% During our experiments, we observed that this type of loss function helps to improve the performance of a shallow network in absence of a large training dataset. 
%Moreover, performance of the proposed models are almost comparable to the various state of the art techniques with regard to various cleanup related tasks. 
%However, for memory and energy constrained devices, the degree of cleanup achieved by a model is not the single most important factor. 
%The size and the associated computational cost of the model should also be considered. 
% Though the state of the art models perform marginally better than our proposed models, the size and computational costs of the existing models make them unsuitable for memory constrained devices. 
%Moreover, the margin by which these existing models perform better than our models may not always be perceptible for the memory constrained devices. 
% Our proposed models give reasonably good performance with the added advantages of lower size and lesser computational cost.

\bibliographystyle{splncs04}
\bibliography{references.bib}

\end{document}